\documentclass{article} % For LaTeX2e
\usepackage{iclr2027_conference,times}

\usepackage{amsmath,amsfonts,bm}

\def\eqref#1{equation~\ref{#1}}
\def\1{\bm{1}}

\DeclareMathAlphabet{\mathsfit}{\encodingdefault}{\sfdefault}{m}{sl}
\SetMathAlphabet{\mathsfit}{bold}{\encodingdefault}{\sfdefault}{bx}{n}

\usepackage{hyperref}
\usepackage{url}

\usepackage[utf8]{inputenc} % allow utf-8 input
\usepackage[T1]{fontenc}    % use 8-bit T1 fonts
\usepackage{hyperref}       % hyperlinks
\usepackage{url}            % simple URL typesetting
\usepackage{booktabs}       % professional-quality tables
\usepackage{amsfonts}       % blackboard math symbols
\usepackage{nicefrac}       % compact symbols for 1/2, etc.
\usepackage{microtype}      % microtypography
\usepackage{xcolor}         % colors
\usepackage{amsmath}
\usepackage{amssymb}
\usepackage{mathtools}
\usepackage{amsthm}
\usepackage{graphicx}
\usepackage{enumitem}
\theoremstyle{definition}
\newtheorem{definition}{Definition}[]
\usepackage{wrapfig}

\newcommand{\real}{\mathbb{R}}

\newcommand{\bx}{\mathbf{x}}

\newcommand{\model}{\textsc{TrajGANR}}
\newcommand{\traj}{r}
\newcommand{\trajdata}{\mathcal{D}^{(\traj)}}
\newcommand{\trajidx}{i}
\newcommand{\numtraj}{N^{(\traj)}}
\newcommand{\trajseq}{\mathbf{r}}
\newcommand{\trajptidx}{j}
\newcommand{\trajseqlen}{L^{(\traj)}}
\newcommand{\trajpt}{\mathbf{p}}
\newcommand{\trajloc}{\bx^{(\traj)}}
\newcommand{\trajtim}{t}

\newcommand{\trajcontlocs}[1]{\mathcal{X}(\trajseq_{#1})}
\newcommand{\trajconttims}[1]{\mathcal{T}(\trajseq_{#1})}

\newcommand{\trajlocenc}{\mathcal{F}_{\traj}^{(x)}}
\newcommand{\trajtimenc}{\mathcal{F}_{\traj}^{(\trajtim)}}

\newcommand{\svi}{I}
\newcommand{\svidata}{\mathcal{D}^{(\svi)}}
\newcommand{\sviidx}{k}
\newcommand{\numsvi}{N^{(\svi)}}   % Changed from previous: Changed N_{I} to N^{(I)} for symmetry with N^{(r)}
\newcommand{\svimg}{\mathbf{o}}
\newcommand{\svimgmat}{\mathbf{I}}
\newcommand{\sviloc}{\bx^{(\svi)}}
\newcommand{\locdata}{\mathcal{D}^{(x)}}

\newcommand{\sext}{\mathcal{S}}
\newcommand{\svienc}{\mathcal{F}^{(\svi)}}
\newcommand{\sviemb}{\mathbf{e}^{(\svi)}}
\newcommand{\svilocenc}{\mathcal{F}_{\svimg}^{(x)}}

\newcommand{\locemb}{\mathbf{e}^{(x)}}

\newcommand{\muldataset}{\mathbb{D}}

\newcommand{\mulgeoembloc}{\mathbf{u}}

\newcommand{\mfsr}{\mathcal{F}}

\newcommand{\trajemb}[1]{\mathbf{e}_{\trajidx}[#1]}
\newcommand{\trajembsviset}[1]{\mathcal{E}_{\traj}(#1)}
\newcommand{\trajembsviagg}{\mathbf{e}^{(\traj)}_{\sviidx}}
\newcommand{\emb}{\mathbf{e}_{\sviidx}}
\newcommand{\trajagg}{\mathcal{G}_{\theta}}
\newcommand{\taskhead}[1]{M_{\psi}^{(#1)}}

\title{\model: \underline{Traj}ectory-Centric Urban \\ Multimodal Learning via \underline{G}eospatially \underline{A}ligned \underline{N}eural \underline{R}epresentations}

\author{%
  Maria Despoina Siampou$^{1,3}$\thanks{Equal contribution},
  Gengchen Mai$^{1,4*}$, Ni Lao$^2$, Jinmeng Rao$^2$, {Neha Arora}$^1$ \\ 
  \textbf{Cyrus Shahabi}$^3$, \textbf{Shushman Choudhury}$^1$ \\
  $^1$Google Research, $^2$Google LLC, $^3$ University of Southern California, $^4$The University of Texas at Austin \\
  \texttt{\{siampou, gengchenm, nlao, jinmengrao, nehaarora\}@google.com} \\
  \texttt{\{shahabi\}@usc.edu} \\
  \texttt{\{shushmac\}@google.com}
}

\iclrfinalcopy % Uncomment for camera-ready version, but NOT for submission.
\begin{document}

\maketitle

\begin{abstract}
Many urban prediction tasks depend not only on the static attributes of a location, such as the visual appearance of its built environment, but also on its dynamic \emph{function}: how people navigate and use it over time. Predicting congestion, travel demand, or road safety risks, for instance, requires mobility information that imagery and geographic coordinates alone cannot provide. Trajectories carry this signal, yet current trajectory representations are difficult to align with static geospatial observations: sparse, irregular GPS samples rarely coincide with street-view image coordinates, and existing encoders produce either a single embedding per trajectory or road segment, or embeddings only at the observed GPS samples. We introduce \textbf{\model{}}, a trajectory-centric multimodal self-supervised learning framework that represents each trajectory as a \emph{continuous, path-conditioned neural field}. Reconstructed from the observed GPS samples, this representation can be queried at arbitrary coordinates (including the off-path locations where street-view images were captured), yielding localized mobility embeddings that align with street-view image and location representations at the same coordinate. Across four fine-grained road-level tasks in San Francisco and Porto, \model{} consistently outperforms recent geospatial and trajectory foundation models, with ablations attributing the gains to both the mobility signal and the fine granularity at which it is aligned. \model{}'s location encoder further generalizes beyond the road network to areas where neither street-view imagery nor trajectories are available.
\end{abstract}

\section{Introduction}

Cities are jointly shaped both by the layout and appearance of their built environment (e.g., streets, buildings, and public spaces) as well as by their patterns of human mobility and activity~\citep{batty2013new}. Many urban properties depend on the latter: road safety risk reflects localized travel behavior~\cite{li2026lagging}, congestion and travel demand reflect collective movement~\citep{zheng2014urban}, and an area's land use is partly revealed by when, how often, and from where people visit it~\citep{yuan2012discovering}. Yet, visual appearance and movement often carry different information: two nearby road segments can look similar in street-view imagery while supporting very different traffic volumes, speeds, or driving behaviors. Static observations, including imagery and coordinates, therefore describe the \emph{form} of urban space, whereas trajectories provide the dynamic signal needed to characterize its \emph{function}.

Learning general-purpose representations of urban space requires integrating these complementary signals~\citep{mai2022review, mai2024opportunities}. Multimodal self-supervised learning (MSSL) offers a natural framework for this integration and has become the main pretraining paradigm in geospatial representation learning~\citep{xiao2025foundation, janowicz2025geofm}. Existing urban MSSL frameworks, however, pretrain almost exclusively on \emph{static} modalities, such as street-view images, satellite imagery, and geographic coordinates, relying on two primary objective families: contrastive approaches that align observations recorded at the same or nearby locations~\citep{cepeda2023geoclip, klemmer2025satclip, mai2023csp, muhlematter2025urbanfusion, liu2026gair, balsebre2023cityfm}, and reconstruction-based approaches that mask and reconstruct multi-source observations represented on a shared spatial grid~\citep{fuller2024croma, guo2024skysense, choudhury2025s2vec, bai2025integrating}. Both families thus presuppose that every modality is observed at a common set of locations, a condition trajectories do not satisfy. Although the value of mobility trajectories for geospatial pretraining has been articulated explicitly~\citep{choudhury2024towards, siampou2025toward}, no existing MSSL framework operationalizes this by aligning GPS trajectories with static urban modalities.

\begin{wrapfigure}{r}{0.42\textwidth}
  \vspace{-\intextsep}
  \centering
  \includegraphics[width=0.42\textwidth]{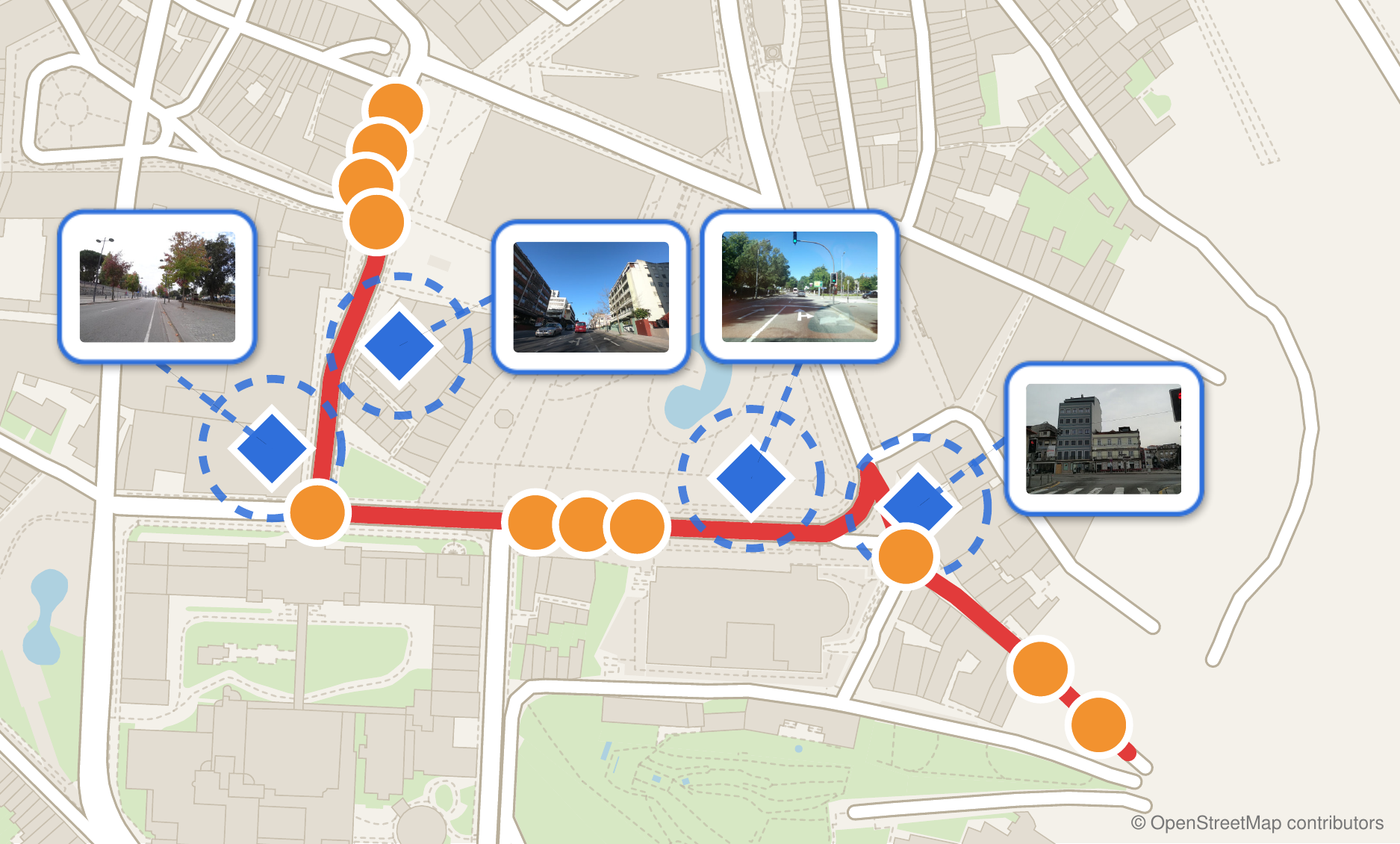}
  \caption{A vehicle's continuous path (red) is observed at irregularly spaced GPS sample waypoints (orange), while street-view images (blue) are captured at locations that typically do not co-locate with waypoints. }
  \label{fig:motivation_fig}
  \vspace{-\intextsep}
\end{wrapfigure}
Integrating trajectories into the pretraining of urban MSSL frameworks is not straightforward. Static modalities such as street-view images (SVIs) are anchored to individual geographic locations, whereas trajectories describe movement along a continuous path through space and time. Figure~\ref{fig:motivation_fig} illustrates this mismatch: a vehicle's movement is observed only at irregularly spaced GPS samples, and nearby street-view images are captured at locations that rarely coincide with them. Existing trajectory encoders cannot directly supply the localized movement signal at an image's coordinate, as they typically produce either a single embedding summarizing the entire trajectory or one embedding per road segment/observed waypoint~\citep{lin2024unite,zhao2025unitr,zhu2025unitraj, lin2024trajfm}. None of these representations can directly answer the location-centered question needed for multimodal learning: \emph{What movement context does this trajectory provide at the coordinate of this image?} Discretizing trajectories into grid cells or map-matched road segments~\citep{li2018deep, ma2024more, wang2025gtr} creates common alignment units, but makes the representation constant within each unit, erasing variation between nearby locations~\citep{zhou2025blurred}. Thus, incorporating trajectories into multimodal pretraining \textbf{requires a representation that preserves continuous variation along a path and can return a localized trajectory embedding at any queried coordinate along or near that path.}

We introduce \textbf{\model{}}, a trajectory-centric geospatial MSSL framework designed around this requirement. Rather than forcing a static observation to match the nearest recorded GPS point or a pooled road-segment representation, \model{} uses the observation's coordinate as a query into the trajectory, which it represents as a continuous, path-conditioned neural field. Given the observed GPS samples together with query coordinates supplied by nearby static modalities (SVI locations in our setup), the model jointly contextualizes the observed and queried positions and infers the trajectory's localized movement context at every query. This produces a localized trajectory embedding at the exact coordinate of each image, even when no GPS point was recorded there. \model{} then aligns that embedding with the corresponding street-view and location representations through a multimodal self-supervised objective. Across many trajectories, this training encourages the resulting location representations to capture both the \emph{form} of urban space and the \emph{function} it supports.

We evaluate \textsc{\model{}} on four road-segment-level downstream tasks in San Francisco and Porto: average traffic speed, road popularity, area-of-interest (AOI) function, and hard-braking event prediction. Together, these tasks isolate the two effects our design targets: the contribution of mobility signals to multimodal geospatial pretraining, and the granularity at which trajectories and static observations are aligned. \textsc{\model{}} outperforms recent geospatial MSSL frameworks that combine imagery with locations, as well as a unimodal trajectory foundation model, on every task. In San Francisco, it reduces traffic-speed error by $44\%$ and hard-braking error by $82\%$ relative to the strongest baseline. Critically, when trajectories are instead aligned using a pooled road-segment representation, this advantage disappears on traffic speed and hard braking prediction tasks.
%, while performance on road popularity and AOI function remains mostly unchanged.  
This result supports the role of fine-grained alignment precisely for outcomes that vary within a segment. Finally, we demonstrate that \textsc{\model{}}'s location encoder generalizes beyond the road network: at off-network locations, where neither street-view imagery nor trajectories are available, it improves F1 by $28\%$ over the strongest baseline for the POI popularity prediction task, demonstrating that mobility-grounded supervision transfers to regions unseen during model pretraining.

In summary, our contributions are:
\begin{itemize}[leftmargin=*,itemsep=1pt, parsep=0pt, topsep=0pt]
\item We introduce \textbf{\textsc{\model}}, a trajectory-centric MSSL framework that aligns human mobility trajectories with street-view images and their geographic locations.
\item We develop a path-conditioned neural implicit function that uses sparse observed GPS points to infer localized movement context at arbitrary query coordinates supplied by static modalities.
\item We develop a fine-grained MSSL objective that aligns each queried trajectory embedding with its street-view and location representations at the same coordinate.
\item We introduce four road-segment-level tasks across two metropolitan areas, on which \textsc{\model{}} outperforms recent geospatial MSSL frameworks and a trajectory foundation model on every task.
% \item We show that these gains come from the mobility signal itself and its alignment granularity.
% \item We further show that \textsc{\model{}}'s location encoder generalizes to locations outside the road network, where neither SVIs nor trajectories are available.
\item Through alignment and modality ablations, we show that the gains arise from both the trajectory signal and the ability to query it at fine spatial granularity.
\item We further show that \textsc{\model{}}'s location encoder generalizes to off-road network locations.
\end{itemize}

\section{Preliminaries}  \label{sec:problem}

\begin{definition}[Street-view Image Dataset]
Let $\sext \subseteq \real^2$ denote the continuous spatial extent of interest, and let $\svidata = \{\svimg_{\sviidx}\}_{\sviidx=1}^{\numsvi}$ be a dataset of $\numsvi$ street-view images georeferenced within $\sext$. Each image is represented as a tuple $\svimg_{\sviidx} = (\svimgmat_{\sviidx}, \sviloc_{\sviidx})$, where $\svimgmat_{\sviidx} \in \real^{C \times H \times W}$ denotes the high-dimensional visual feature tensor and $\sviloc_{\sviidx} \in \sext$ is its corresponding geographic coordinate. We denote the set of image locations by $\locdata = \{\sviloc_{\sviidx}\}_{\sviidx=1}^{\numsvi} \subset \sext$. In our study, $\sext$ corresponds to a bounded region.
\end{definition}

\begin{definition}[GPS Trajectory Dataset]
Let $\trajdata = \{\trajseq_{\trajidx}\}_{\trajidx=1}^{\numtraj}$ be a collection of $\numtraj$ GPS trajectories observed within $\sext$ over a temporal horizon $\mathcal{T} \subseteq \real$. Each trajectory $\trajseq_{\trajidx}$ is an ordered sequence of $\trajseqlen_{\trajidx}$ sampled GPS points, $\trajseq_{\trajidx} = (\trajpt_{\trajidx 1}, \trajpt_{\trajidx 2}, \ldots, \trajpt_{\trajidx \trajseqlen_{\trajidx}})$, where each $\trajpt_{\trajidx\trajptidx} = (\trajloc_{\trajidx\trajptidx}, \trajtim_{\trajidx\trajptidx})$ consists of a location $\trajloc_{\trajidx\trajptidx} \in \sext$ and a strictly increasing timestamp $\trajtim_{\trajidx\trajptidx} \in \mathcal{T}$ such that $\trajtim_{\trajidx 1} < \trajtim_{\trajidx 2} < \cdots < \trajtim_{\trajidx \trajseqlen_{\trajidx}}$. Although observed as a finite sequence of discrete points, $\trajseq_{\trajidx}$ samples an underlying continuous movement trajectory through space and time. We denote this continuous spatial footprint and temporal span by $\trajcontlocs{\trajidx} \subset \sext$ and $\trajconttims{\trajidx} \subseteq \mathcal{T}$, respectively, such that $\trajloc_{\trajidx\trajptidx} \in \trajcontlocs{\trajidx}$ and $\trajtim_{\trajidx\trajptidx} \in \trajconttims{\trajidx} \forall \trajptidx \in \{1,\ldots,\trajseqlen_{\trajidx}\}$.
\end{definition}

\textbf{Problem Formulation.} Let $\muldataset = (\svidata, \trajdata)$ denote a multimodal geospatial dataset defined over the continuous spatial domain $\sext$. Our objective is to pretrain a parameterized representation function $\mfsr_\theta$ on $\muldataset$. For any geographic query coordinate $\bx \in \sext$, the function produces a continuous localized neural embedding $\mulgeoembloc(\bx) = \mfsr_\theta(\bx; \muldataset) \in \real^{d},$ where $\mulgeoembloc(\bx)$ encodes localized multimodal context by integrating \textit{visual features} from $\svidata$ and \textit{mobility patterns} from $\trajdata$ within the neighborhood of $\bx$. 
%This pretrained representation function can subsequently be fine-tuned for downstream road and urban-level understanding tasks.

\section{Methodology}  \label{sec:method}

\begin{figure*}
    \centering
    %\vspace{-13mm}
    \includegraphics[width=\linewidth]{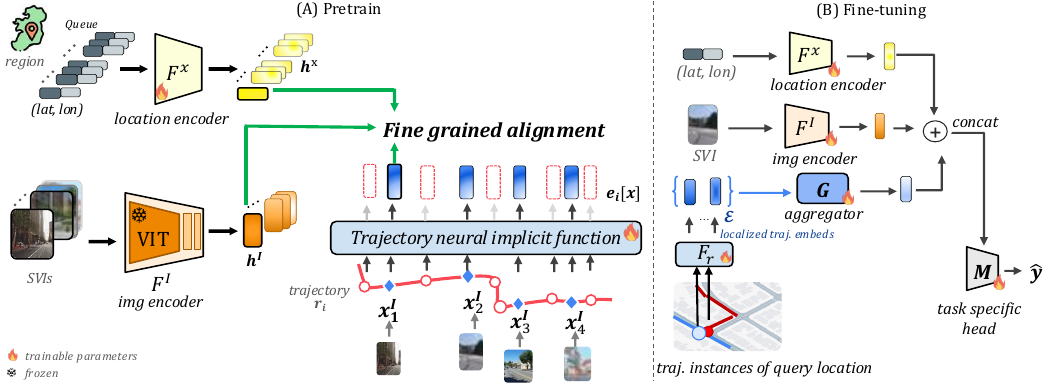}
    \caption{\textbf{Overview of \model{}.} \textbf{(a)~Pretraining:} Fine-grained geospatial alignment jointly trains trajectories, street-view images, and coordinates. White circles denote observed points  of $\trajseq_{\trajidx}$, while blue diamonds indicate SVI locations $\{\sviloc_{\sviidx}\}$. \textbf{(b)~Fine-Tuning:} For a query location $\sviloc_{\sviidx}$, localized trajectory embeddings $\trajembsviset{\sviloc_{\sviidx}}$ are queried, aggregated, and concatenated with image ($\sviemb_{\sviidx}$) and location ($\locemb_{\sviidx}$) representations for downstream prediction.}
    \label{fig:overview}
    %\vspace{-5mm}
\end{figure*}

In this section, we present the components of our {\model} framework, as illustrated in Fig.~\ref{fig:overview}.

\subsection{Static Modality Encoders} \label{sec:encoder}

We first describe the encoding of the \textbf{static} modalities of $\muldataset$, which include the street-view image dataset $\svidata$ and their geographic coordinate set $\locdata$. Following standard practices in geospatial multimodal representation learning~\citep{cepeda2023geoclip, liu2026gair}, we encode them using specialized, modality-specific encoders to project them into a shared $d$-dimensional embedding space.

\textbf{Street-view Image Encoder.} 
Let $\svienc$ denote the street-view image encoder. Given the image tensor $\svimgmat_{\sviidx} \in \real^{C \times H \times W}$ of tuple $\svimg_{\sviidx} = (\svimgmat_{\sviidx}, \sviloc_{\sviidx})$, the encoder produces a visual representation $\sviemb_{\sviidx} \in \real^{d}$:
\begin{equation}
    \sviemb_{\sviidx} = \svienc(\svimgmat_{\sviidx}),
\end{equation}
where $\svienc = g_{\mathrm{I}} \circ \phi_{\mathrm{I}}$. Here, $\phi_{\mathrm{I}}$ is a frozen, pretrained Vision Transformer (ViT) backbone~\citep{dosovitskiy2020image} while $g_{\mathrm{I}}$ is a learnable projection head that adapts the visual features to a shared embedding space. We keep the $\phi_{\mathrm{I}}$ frozen to prevent its generalized representations from overfitting to our localized spatial distribution, and minimize computational overhead during cross-modal training.

\textbf{Street-view Location Encoder.}
Let $\svilocenc$ denote the geographic location encoder. Given the image coordinates $\sviloc_{\sviidx} \in \sext$ of street-view image $\svimg_{\sviidx}$, the encoder produces a location embedding $\locemb_{\sviidx} \in \real^{d}$:
\begin{equation}
    \locemb_{\sviidx} = \svilocenc(\sviloc_{\sviidx}),
\end{equation}
where $\svilocenc$ is parameterized using Space2Vec~\citep{mai2020space2vec}. We choose Space2Vec as it is specifically designed to preserve multi-scale spatial frequency relationships within a local study area.\footnote{For global scale modeling, alternative encoding approaches such as Random Fourier Features~\citep{cepeda2023geoclip} could be utilized. Similarly, to encode more complex, non-point geographic geometries, methods like Poly2Vec~\citep{siampou2025poly2vec} represent orthogonal approaches that can be integrated into this framework.}

\subsection{Trajectory Neural Implicit Function} \label{sec:traj_nif}

We now describe our modeling approach for the \textbf{dynamic} modality $\trajdata$ of $\muldataset$. Aligning trajectories with static modalities in real-world scenarios is fundamentally challenged by a spatial mismatch: trajectories represent continuous movement, while their sparse and irregularly sampled GPS points rarely co-locate with street-view camera locations. We resolve this challenge by formulating trajectories \textbf{as continuous neural implicit representations} that can be queried at image's location coordinates. This allows {\model} to infer a localized trajectory embedding at the exact capture site of any nearby street-view image, enabling cross-modal alignment at the most granular level. 

In the following, we detail the encoding of discrete GPS observations (Section~\ref{sec:point_tokenization}), formalize the coordinate-conditioned query mechanism (Section~\ref{sec:query_encoding}), and present the joint spatiotemporal contextualization that reconstructs the continuous coordinate-conditioned trajectory field (Section~\ref{sec:traj_transformer}).

\subsubsection{Trajectory Observed Points Encoding} \label{sec:point_tokenization}

Each trajectory $\traj_{\trajidx} \in \trajdata$ is represented as a sequence of discretely sampled GPS points $\trajpt_{\trajidx\trajptidx}$, which we refer to as observed points. Each such point is characterized by a pair of coordinates $\trajloc_{\trajidx\trajptidx} \in \sext$ and a timestamp $\trajtim_{\trajidx\trajptidx} \in \mathcal{T}$, which we project into a shared $d$-dimensional space via two separate factorized encoders. We then combine them additively to construct our initial trajectory tokens:
\begin{equation}
    \mathbf{h}_{\trajidx\trajptidx}^{0} = \trajlocenc(\trajloc_{\trajidx\trajptidx}) + \trajtimenc(\trajtim_{\trajidx\trajptidx}).
\end{equation}
Here, $\trajlocenc: \sext \to \real^d$ is a trajectory-specific Space2Vec location encoder~\citep{mai2020space2vec} and $\trajtimenc: \real \to \real^d$ is a Time2Vec temporal encoder~\citep{kazemi2019time2vec}. By summing these two representations, the temporal encoding $\trajtimenc(\trajtim_{\trajidx\trajptidx})$ served as a positional encoding, structurally preserving the sequential order of the points along the trajectory.

\subsubsection{Coordinate-Conditioned Query Construction} \label{sec:query_encoding}

To align trajectories with static modalities at the most granular geographic level, we treat each street-view image's coordinate as a spatial query that probes the continuous trajectory footprint. 

Formally, for a given trajectory $\traj_{\trajidx}$, let $\mathcal{Q}_{\trajidx} = \{\bx \in \sext \mid d(\bx, \trajcontlocs{\trajidx}) \le \epsilon\}$\footnote{In practice, $\epsilon$ is set to a 20-50\,m buffer on the order of typical GPS error to make sure that the street-view images associated with query locations were taken along the road traversed by trajectories while we filter out SVIs that are taken in completely off-road areas (e.g., parking lots).} denote a set of queries representing street-view image locations $\sviloc_{\sviidx}$ within a distance threshold $\epsilon$ of the trajectory's continuous spatial footprint $\trajcontlocs{\trajidx}$. For each query coordinate $\bx \in \mathcal{Q}_{\trajidx}$, we construct an initial query token:
\begin{equation}
    \mathbf{q}_{\trajidx}^{0}[\bx] = \trajlocenc(\bx) + \mathbf{t}_{\trajidx}[\bx],
\end{equation}
where $\trajlocenc(\bx)$ provides spatial grounding using the trajectory-specific location encoder, and $\mathbf{t}_{\trajidx}[\bx] \in \real^d$ represents the trajectory-conditioned temporal embedding assigned to $\bx$. We clarify that $\trajlocenc(\bx)$ is distinct from the street-view location encoder $\svilocenc(\bx)$ defined in Section~\ref{sec:encoder}, as the latter is intended to provide general regional geographic context, whereas $\trajlocenc(\bx)$ aims to project coordinates into a subspace that aligns with sequential and directional dynamics of trajectories.

We now describe how the trajectory-conditioned temporal embedding $\mathbf{t}_{\trajidx}[\bx] \in \real^d$ is computed. Given that street-view images are captured independently of the trajectory dataset, our query set $\mathcal{Q}_{\trajidx}$ lacks intrinsic timestamps. While image capture times could be used, we choose to not constrain our framework, since our goal is not to match transient, real-time events (e.g., co-locating a vehicle with a simultaneous GPS ping) but to learn the \textbf{long-term visual and behavioral characteristics of a location}. Since the physical infrastructure of a neighborhood (e.g., road layouts, neighborhood aesthetics) evolves slowly, asynchronous images still accurately reflect the environment that dictates current trajectory patterns. Under this assumption, we investigate two query-time encoding strategies:

\paragraph{Synthetic-Time Query Encoding.} The first strategy assigns a synthetic timestamp to the query location. Since $\trajseq_{\trajidx}$ represents continuous movement over the spatial footprint $\trajcontlocs{\trajidx}$ and temporal span $\trajconttims{\trajidx}$, a query location $\bx \in \trajcontlocs{\trajidx}$ can be associated with an estimated time $\hat{\trajtim}_{\trajidx}[\bx] \in \trajconttims{\trajidx}$. Let $\trajpt_{\trajidx a}$ and $\trajpt_{\trajidx b}$ denote two neighboring observed GPS samples along $\trajseq_{\trajidx}$ such that $\bx$ lies between their locations along the trajectory footprint. We estimate the query time by linear interpolation:
\begin{equation}
    \hat{\trajtim}_{\trajidx}[\bx] = (1-\lambda)\trajtim_{\trajidx a} + \lambda \trajtim_{\trajidx b},
\end{equation}
where $\lambda \in [0,1]$ denotes the relative position of $\bx$ between $\trajloc_{\trajidx a}$ and $\trajloc_{\trajidx b}$ along the trajectory, with
$\lambda = \frac{\|\bx - \trajloc_{\trajidx a}\|_2}{\|\trajloc_{\trajidx b} - \trajloc_{\trajidx a}\|_2}.$
The temporal embedding is then obtained using the temporal encoder:
\begin{equation}
    \mathbf{t}_{\trajidx}[\bx] = \trajtimenc\left(\hat{\trajtim}_{\trajidx}[\bx]\right).
\end{equation}
This variant treats the query location as a point that the trajectory traverses at an estimated time.

%\maria{TODO: Look at masked time encoding again. Not sure if this description is accurate.}
\paragraph{Masked-Time Query Encoding.} The second strategy represents time as a latent property by assigning a sequence-level learnable mask to each query. Specifically, for a query set of size $k$, we assign a sequence of learnable temporal masks $\{\mathbf{m}_{\mathrm{time}}^{(j)}\}_{j=1}^k$ such that:
\begin{equation}
    \mathbf{t}_{\trajidx}[\bx_j] = \mathbf{m}_{\mathrm{time}}^{(j)},
\end{equation}
where $\mathbf{m}_{\mathrm{time}}^{(j)} \in \mathbb{R}^{d}$ is optimized end-to-end. This approach treats each query's temporal feature as a mask, allowing the self-attention mechanism to dynamically reconstruct the missing timestamp of $\bx_j$ by contextualizing its unique spatial encoding $\trajlocenc(\bx_j)$ against the observed trajectory points.

\subsubsection{Joint Spatiotemporal Contextualization} \label{sec:traj_transformer}

To obtain localized trajectory embeddings at the query locations, we jointly process the observed trajectory sequence and the queries. Recall from Section~\ref{sec:query_encoding} that the query set $\mathcal{Q}_{\trajidx}$ contains the coordinates of all street-view images located in the neighborhood of trajectory's $\trajseq_{\trajidx}$ continuous spatial footprint $\trajcontlocs{\trajidx}$. To preserve this relationship while modeling the path's dynamics, we concatenate the sequence of initial trajectory tokens $\mathbf{H}_{\trajidx}^{0} = [\mathbf{h}_{\trajidx 1}^{0}, \ldots, \mathbf{h}_{\trajidx \trajseqlen_{\trajidx}}^{0}]$ of $\trajcontlocs{\trajidx}$ with its set of spatial query tokens $\{\mathbf{q}_{\trajidx}^{0}[\bx]\}_{\bx \in \mathcal{Q}_{\trajidx}}$. This yields an augmented spatiotemporal sequence:
\begin{equation}
    \widetilde{\mathbf{H}}_{\trajidx}^{0} = \left[ \mathbf{h}_{\trajidx 1}^{0}, \ldots, \mathbf{h}_{\trajidx \trajseqlen_{\trajidx}}^{0}, \big\{\mathbf{q}_{\trajidx}^{0}[\bx]\big\}_{\bx \in \mathcal{Q}_{\trajidx}} \right].
\end{equation}
We pass the augmented sequence to an $L$-layer Transformer encoder backbone~\citep{vaswani2017attention}:
\begin{equation}
    \widetilde{\mathbf{H}}_{\trajidx}^{L} = \mathrm{Transformer}\left(\widetilde{\mathbf{H}}_{\trajidx}^{0}\right).
\end{equation}
Through self-attention, each query token is processed in parallel with the  observed sequence, allowing the Transformer to implicitly interpolate the trajectory's movement patterns at each continuous query location $\bx$. The final-layer representation of each $\mathbf{q}_{\trajidx}^{0}[\bx]$ serves as the localized trajectory embedding:
\begin{equation}
     \trajemb{\bx} = \widetilde{\mathbf{H}}_{\trajidx,\mathrm{idx}(\bx)}^{L} \in \real^d,
\end{equation}
where $\mathrm{idx}(\bx)$ maps the query coordinate to its corresponding positional index in the augmented sequence $\widetilde{\mathbf{H}}_{\trajidx}^{L}$. The representation  $\trajemb{\bx}$ captures the localized movement context of trajectory $\traj_{\trajidx}$ at coordinate $\bx$, and is subsequently aligned with the visual and location representations of the corresponding street-view image. We describe this alignment process in the next section.

\subsection{\model{} Multimodal Self-Supervised Pretraining}
\label{sec:pretraining}

For each street-view coordinate $\sviloc_{\sviidx} \in \mathcal{Q}_{\trajidx}$, the framework yields three representations co-anchored within the same local neighborhood: the visual feature $\sviemb_{\sviidx}$, the global location prior $\locemb_{\sviidx}$, and the localized trajectory embedding $\trajemb{\sviloc_{\sviidx}}$. We denote by $\mathcal{L}_{\mathrm{info}}(\mathbf{u}, \mathbf{v}^+; \mathcal{N})$ the standard InfoNCE loss~\citep{oord2018representation} between an anchor $\mathbf{u}$ and a positive target $\mathbf{v}^+$, contrasted against a negative set $\mathcal{N}$.
%under cosine similarity with temperature $\tau$. 
The overall pretraining objective is the sum of all cross-modal pairs:
\begin{equation}
    \mathcal{L}_{\mathrm{pre}} = \underbrace{\mathcal{L}_{\mathrm{info}}(\sviemb_{\sviidx}, \locemb_{\sviidx}; \mathcal{N}_{\mathrm{loc}})}_{\text{Image--Location}} + \underbrace{\mathcal{L}_{\mathrm{info}}(\trajemb{\sviloc_{\sviidx}}, \locemb_{\sviidx}; \mathcal{N}_{\mathrm{loc}})}_{\text{Trajectory--Location}} + \underbrace{\mathcal{L}_{\mathrm{info}}(\trajemb{\sviloc_{\sviidx}}, \sviemb_{\sviidx}; \mathcal{N}_{\mathrm{img}})}_{\text{Trajectory--Image}}.
    \label{eq:pretraining_loss}
\end{equation}
Here, the negative set $\mathcal{N}_{\mathrm{loc}}$ consists of coordinates sampled from a MoCo-style memory queue~\citep{he2020momentum}, while $\mathcal{N}_{\mathrm{img}}$ draws in-batch negatives to contrast the positive street-view against images of different locations. Furthermore, by design, a single street-view coordinate $\sviloc_{\sviidx}$ can have multiple trajectories passing through its local neighborhood. Each of these trajectories generates a unique localized embedding $\trajemb{\sviloc_{\sviidx}}$, providing multiple positive trajectory views for the same static anchor $\sviloc_{\sviidx}$. This multi-view contrastive setup incentivizes {\model} to extract generalized, location-consistent mobility patterns rather than overfitting to or memorizing individual trajectory instances.

\subsection{\model{} Fine-Tuning} \label{sec:finetuning}

To adapt our pretrained encoders to downstream prediction tasks, we construct a single multimodal final embedding $\emb$ for each target location $\sviloc_{\sviidx}$. Each place is usually associated with a single image $\sviemb_{\sviidx}$ and location embeddings $\locemb_{\sviidx}$, but a variable-sized set of localized trajectory embeddings $\trajembsviset{\sviloc_{\sviidx}} = \{ \trajemb{\sviloc_{\sviidx}} \mid \sviloc_{\sviidx} \in \mathcal{Q}_{\trajidx} \}$. To aggregate this set into a single vector of fixed dimensionality, we feed the set into a permutation-invariant multi-head self-attention pooling block $\trajagg$ prepended with a learnable $\texttt{[CLS]}$ token. The final-layer $\texttt{[CLS]}$ embedding is the collective mobility profile:
\begin{equation}
    \trajembsviagg = \trajagg\left( \trajembsviset{\sviloc_{\sviidx}} \right). \label{eq:traj_agg}
\end{equation}
We then concatenate these three modalities to get the final location representation: $\emb = \left[\sviemb_\sviidx \parallel \locemb_\sviidx \parallel \trajembsviagg \right]$, where $\parallel$ denotes concatenation. Finally, for each downstream task $k$, we attach a task-specific prediction head $\taskhead{k}$ to generate the target prediction, which is optimized end-to-end using the corresponding task's supervised loss:
% \begin{equation}
%     \emb = \left[\sviemb_\sviidx \parallel \locemb_\sviidx \parallel \trajembsviagg \right], \label{eq:loc_agg}
% \end{equation}
% where $\parallel$ denotes concatenation. Finally, for each downstream task $k$, we attach a task-specific prediction head $\taskhead{k}$ to generate the target prediction:
\begin{equation}
    \hat{\mathbf{y}}_{\sviidx}^{(k)} = \taskhead{k}(\emb).
\end{equation}
% which is optimized end-to-end using the corresponding task's supervised loss.

\section{Experiments}\label{sec:exp}

\subsection{Experimental Setup} 

\textbf{Datasets.} We evaluate \model{} using two publicly available and widely adopted trajectory datasets: the Porto Taxi dataset~\citep{o2015ecml} and the San Francisco Cabspotting dataset~\footnote{https://stamen.com/work/cabspotting/}. Following previous works, we download street-view images and their capture locations from the Mapillary API~\footnote{https://www.mapillary.com/}~\citep{liu2026gair}. Additional dataset details can be found in Appendix~\ref{apdx:dataset-stats}.

\textbf{Baselines.} %\model{} is the first framework to align static geospatial modalities with dynamic trajectory data to learn location embeddings. 
We compare \model{} against representative baselines across two complementary categories: (i)~\textbf{static geospatial foundation models}, including \textsc{GeoCLIP}~\citep{cepeda2023geoclip}, \textsc{SatCLIP}~\citep{klemmer2025satclip}, and \textsc{GAIR}~\citep{liu2026gair}, and (ii)~\textbf{trajectory-only foundation model}, including the state-of-the-art \textsc{UniTraj}~\citep{zhu2025unitraj} to assess whether mobility dynamics alone suffice for downstream tasks. During fine-tuning, we freeze all pretrained encoders and train a lightweight, task-specific prediction head. For \textsc{UniTraj}, we mean-pool the point embeddings along the road segment corresponding to each target location. More details are provided in Appendix~\ref{apdx:baselines}.

\textbf{Evaluation Tasks.} We evaluate \model{} on four downstream fine-grained road-level tasks derived from Google Maps: (i)~\textbf{traffic speed} prediction, which estimates the historical average driving speed along each road segment, (ii)~\textbf{road popularity}, which categorizes segments based on aggregated user navigation demand, (iii)~\textbf{AOI function} prediction, which predicts the distribution of Point-of-Interest (POI) categories within the Area of Interest (AOI) surrounding each segment, and 
(iv)~\textbf{hard braking events (HBE)} prediction, which estimates the frequency of abrupt deceleration events as a proxy for driving difficulty and safety risk~\citep{li2026lagging}. Task ground-truth labels are anchored to street-view image coordinates, which serve as the target locations. Due to sparse coverage in Porto, the traffic speed estimation tasks is skipped. We provide more details in Appendix~\ref{apdx:downstream_tasks}.

\subsection{Main Results}

\begin{table*}[t]
\centering
\caption{Main experimental results on San Francisco. \textbf{Best} and \underline{second best} values are highlighted. Results are averaged over 3 runs with different train/test splits. "${\scriptscriptstyle \pm X}$" denotes the standard deviation.}
\label{tab:sf-results}
\footnotesize
\setlength{\tabcolsep}{2pt} % Re-introduce some spacing for readability
\resizebox{0.8\textwidth}{!}{% This forces the table to fit the page width perfectly
\begin{tabular}{l cc cc cc cc}
\toprule
\textbf{Method} & \multicolumn{2}{c}{\textit{traffic speed}} & \multicolumn{2}{c}{\textit{road popularity}} & \multicolumn{2}{c}{\textit{AOI function}} & \multicolumn{2}{c}{\textit{HBE}} \\
\cmidrule(lr){2-3} \cmidrule(lr){4-5} \cmidrule(lr){6-7} \cmidrule(lr){8-9}
& ($\downarrow$) MAE & ($\downarrow$) RMSE & ($\uparrow$) F1 & ($\uparrow$) AUPRC & ($\downarrow$) L1 & ($\uparrow$) Cosine & ($\downarrow$) MAE & ($\downarrow$) RMSE \\
\midrule
\textsc{UniTraj} 
    & $2.353{\scriptscriptstyle \pm.06}$ & $5.303{\scriptscriptstyle \pm.08}$ 
    & $0.181{\scriptscriptstyle \pm.01}$ & $0.354{\scriptscriptstyle \pm.05}$ 
    & $1.194{\scriptscriptstyle \pm.01}$ & $0.602{\scriptscriptstyle \pm.00}$ 
    & $2.373{\scriptscriptstyle \pm.01}$ & $5.975{\scriptscriptstyle \pm.01}$ \\
\midrule
\textsc{GeoCLIP} 
    & $2.607{\scriptscriptstyle \pm.04}$ & $4.848{\scriptscriptstyle \pm.12}$ 
    & $0.431{\scriptscriptstyle \pm.01}$ & $0.615{\scriptscriptstyle \pm.20}$ 
    & $1.144{\scriptscriptstyle \pm.05}$ & $0.634{\scriptscriptstyle \pm.03}$ 
    & $5.853{\scriptscriptstyle \pm.27}$ & $7.297{\scriptscriptstyle \pm.35}$ \\

\textsc{SatCLIP} 
    & $2.316{\scriptscriptstyle \pm.24}$ & $5.239{\scriptscriptstyle \pm.09}$ 
    & $0.181{\scriptscriptstyle \pm.01}$ & $0.393{\scriptscriptstyle \pm.06}$ 
    & $1.202{\scriptscriptstyle \pm.00}$ & $0.600{\scriptscriptstyle \pm.00}$ 
    & $10.95{\scriptscriptstyle \pm.46}$ & $13.11{\scriptscriptstyle \pm.09}$ \\
\textsc{GAIR$_{\text{ SVI}}$}    
    & $2.203{\scriptscriptstyle \pm.09}$ & $4.384{\scriptscriptstyle \pm.04}$
    & $0.582{\scriptscriptstyle \pm.06}$ & {$0.894{\scriptscriptstyle \pm.02}$}  
    & $1.114{\scriptscriptstyle \pm.01}$ & $0.649{\scriptscriptstyle \pm.00}$   
    & $1.744{\scriptscriptstyle \pm.04}$ & $5.061{\scriptscriptstyle \pm.11}$ \\

\textsc{GAIR$_{\text{ loc}}$}    
    & {$1.420{\scriptscriptstyle \pm.18}$} & {$2.981{\scriptscriptstyle \pm.09}$} 
    & {$0.748{\scriptscriptstyle \pm.03}$} & {$0.832{\scriptscriptstyle \pm.02}$}  
    & {$0.570{\scriptscriptstyle \pm.03}$} & {$0.879{\scriptscriptstyle \pm.01}$} 
    & {$1.185{\scriptscriptstyle \pm.24}$} & {$3.586{\scriptscriptstyle \pm.56}$} \\

\textsc{GAIR$_{\text{ loc + SVI}}$}    
    & \underline{$0.725{\scriptscriptstyle \pm.03}$} & \underline{$1.694{\scriptscriptstyle \pm.01}$} 
    & \underline{$0.930{\scriptscriptstyle \pm.02}$} & \underline{$0.981{\scriptscriptstyle \pm.00}$}  
    & \underline{$0.262{\scriptscriptstyle \pm.01}$} & \underline{$0.958{\scriptscriptstyle \pm.00}$} 
    & \underline{$0.715{\scriptscriptstyle \pm.07}$} & \underline{$1.742{\scriptscriptstyle \pm.08}$} 
    \\
\midrule
\textbf{TrajGANR} 
    & \textbf{0.403${\scriptscriptstyle \pm.12}$} & \textbf{1.187${\scriptscriptstyle \pm.22}$} 
    & \textbf{0.960${\scriptscriptstyle \pm.02}$} & \textbf{0.992${\scriptscriptstyle \pm.00}$} 
    & \textbf{0.121${\scriptscriptstyle \pm.03}$} & \textbf{0.981${\scriptscriptstyle \pm.01}$} 
    & \textbf{0.130${\scriptscriptstyle \pm.02}$} & \textbf{0.926${\scriptscriptstyle \pm.19}$} \\
\bottomrule
\end{tabular}
}
\end{table*}
\begin{table*}[t]
\centering
\vspace{-2mm}
\caption{Main experimental results on Porto. \textbf{Best} and \underline{second best} values are highlighted.}
\label{tab:porto-results}
\footnotesize
\setlength{\tabcolsep}{4pt} % Re-introduce some spacing for readability
\resizebox{0.7\textwidth}{!}{% This forces the table to fit the page width perfectly
\begin{tabular}{l cc cc cc}
\toprule
\textbf{Method} & \multicolumn{2}{c}{\textit{road popularity}} & \multicolumn{2}{c}{\textit{AOI function}} & \multicolumn{2}{c}{\textit{HBE}} \\
\cmidrule(lr){2-3} \cmidrule(lr){4-5} \cmidrule(lr){6-7}
& ($\uparrow$) F1 & ($\uparrow$) AUPRC & ($\downarrow$) L1 & ($\uparrow$) Cosine & ($\downarrow$) MAE & ($\downarrow$) RMSE \\
\midrule
\textsc{UniTraj}
    & ${0.383\scriptscriptstyle \pm.05}$ & ${0.459\scriptscriptstyle \pm.08}$
    & {${0.865\scriptscriptstyle \pm.01}$} & ${0.728\scriptscriptstyle \pm.00}$
    & ${1.121\scriptscriptstyle \pm.13}$ & ${4.378\scriptscriptstyle \pm.31}$ \\
\midrule
\textsc{GeoCLIP}
    & ${0.632\scriptscriptstyle \pm.04}$ & ${0.685\scriptscriptstyle \pm.01}$
    & ${0.777\scriptscriptstyle \pm.02}$ & ${0.778\scriptscriptstyle \pm.01}$
    & ${9.807\scriptscriptstyle \pm 1.2}$ & ${14.39\scriptscriptstyle \pm1.2}$ \\
\textsc{SatCLIP}
    & ${0.153\scriptscriptstyle \pm.01}$ & ${0.373\scriptscriptstyle \pm.08}$
    & ${0.856\scriptscriptstyle \pm.03}$ & ${0.737\scriptscriptstyle \pm.01}$
    & ${9.971\scriptscriptstyle \pm 1.3}$ & ${14.80\scriptscriptstyle \pm 1.4}$ \\
\textsc{GAIR$_{\text{ SVI}}$}
    & ${0.449\scriptscriptstyle \pm.26}$ & ${0.557\scriptscriptstyle \pm.24}$
    & ${0.758\scriptscriptstyle \pm.02}$ & {${0.784\scriptscriptstyle \pm.01}$}
    & {${6.180\scriptscriptstyle \pm.10}$} & {${8.842\scriptscriptstyle \pm.08}$} \\
\textsc{GAIR$_{\text{ loc}}$}
    & {${0.795\scriptscriptstyle \pm.02}$} & {${0.823\scriptscriptstyle \pm.02}$}
    & ${0.644\scriptscriptstyle \pm.01}$ & {${0.837\scriptscriptstyle \pm.00}$}
    & ${1.081\scriptscriptstyle \pm.13}$ & ${3.923\scriptscriptstyle \pm.38}$ \\
\textsc{GAIR$_{\text{ loc + SVI}}$}
    & \underline{${0.894\scriptscriptstyle \pm.04}$} & \underline{${0.910\scriptscriptstyle \pm.06}$}
    & $\underline{0.496\scriptscriptstyle \pm.12}$ & \underline{${0.891\scriptscriptstyle \pm.03}$}
    & $\underline{0.937\scriptscriptstyle \pm.04}$ & $\underline{2.652\scriptscriptstyle \pm.16}$ \\
\midrule
\textbf{TrajGANR}
    & \textbf{0.908${\scriptscriptstyle \pm.04}$} & \textbf{0.939${\scriptscriptstyle \pm.04}$}
    & \textbf{0.279${\scriptscriptstyle \pm.01}$} & \textbf{0.951${\scriptscriptstyle \pm.00}$}
    & \textbf{0.401${\scriptscriptstyle \pm.28}$} & \textbf{1.358${\scriptscriptstyle \pm.04}$} \\
\bottomrule
\end{tabular}
}
%\vspace{-3mm}
\end{table*}

Tables~\ref{tab:sf-results} and~\ref{tab:porto-results} report downstream evaluation results across San Francisco and Porto, respectively. Across both regions and all tasks, \model{} consistently outperforms all baseline models. The gains are most pronounced on tasks that depend on driving behavior, namely traffic speed estimation and hard-braking event (HBE) prediction. On these tasks, \model{} reduces MAE over the strongest baseline by $44.4\%$ and $81.8\%$ in San Francisco, and by $57.2\%$ on HBEs in Porto. Among static baselines, \textsc{GAIR}$_{\mathrm{Loc+SVI}}$ is the most competitive, indicating that geographic coordinates combined with street-level visual features provide a strong structural prior. In contrast, while satellite-trained models like \textsc{SatCLIP} can capture broad regional land-use patterns, their overhead perspective misses the localized ground-level details essential for our fine-grained tasks. Finally, we observe that \textsc{UniTraj} struggles in performance despite being pretrained on large-scale trajectory data. We attribute this to two factors: First, \textsc{UniTraj} outputs embeddings strictly at the observed waypoints, lacking a continuous mechanism to infer representations at unseen target coordinates. Thus, it requires segment-level pooling, which averages out fine-grained movement. Second, as a unimodal model, it lacks cross-modal alignment with other modalities, preventing it from grounding its representations in the physical built environment. Overall, these findings demonstrate that \model{}'s advantage comes from integrating complementary static visual features with dynamic movement patterns, as well as its continuous implicit function mechanism that queries trajectory embeddings directly at any location, bypassing the need for lossy segment pooling.
%We isolate these two factors in Sections~\ref{sec:modality-ablation} and~\ref{sec:granularity-ablation}.
 
\subsection{Ablation Studies}

\textbf{The Effect of Alignment Granularity.} We examine how spatial alignment granularity affects representation quality across three pretraining variants: (1)~\textsc{No Align}, which skips multimodal contrastive alignment entirely, (2)~\textsc{Segment-Based}, which aligns modalities at the road-segment level by mean-pooling observed trajectory point embeddings, and (3)~\textsc{Fine-Grain} (our full model), which aligns modalities directly at street-view coordinates using our continuous neural implicit function. As shown in Table~\ref{tab:align_ablation}, \textsc{No Align} yields the weakest performance, while \textsc{Fine-Grain} achieves the strongest overall results, with large gains over \textsc{Segment-Based} on tasks driven by driving behavior, such as traffic speed (MAE reduced by 43\%) and HBE prediction (MAE reduced by 60\%). Segment pooling assigns the same trajectory representation to every SVI on a segment, averaging away how driving behavior varies along it, whereas fine-grained alignment learns its own trajectory-conditioned signal, yielding a richer description of how the road is driven. In contrast, \textsc{Segment-Based} performs comparably on AOI function and road popularity, which depend less on driving behavior. Fine-grained alignment is thus preferable in practice: it matches segment pooling where driving behavior matters less, and substantially outperforms it where it matters most.

%Pretraining at the finest granularity is thus a better design choice, as point-level representations can be aggregated upward when a coarser view is needed, while the converse is not possible.

\begin{table*}[!ht]
\centering
\vspace{-5mm}
\caption{Ablation of alignment granularity on San Francisco. \textbf{Best} values are highlighted.}
\label{tab:align_ablation}
\footnotesize
\setlength{\tabcolsep}{2pt} % Re-introduce some spacing for readability
\resizebox{0.85\textwidth}{!}{% This forces the table to fit the page width perfectly
\begin{tabular}{l cc cc cc cc}
\toprule
\textbf{Method} & \multicolumn{2}{c}{\textit{Traffic Speed}} & \multicolumn{2}{c}{\textit{Road Popularity}} & \multicolumn{2}{c}{\textit{AOI Function}} & \multicolumn{2}{c}{\textit{HBE}} \\
\cmidrule(lr){2-3} \cmidrule(lr){4-5} \cmidrule(lr){6-7} \cmidrule(lr){8-9}
& ($\downarrow$) MAE & ($\downarrow$) RMSE & ($\uparrow$) F1 & ($\uparrow$) AUPRC & ($\downarrow$) L1 & ($\uparrow$) Cosine & ($\downarrow$) MAE & ($\downarrow$) RMSE \\
\midrule
\textsc{No align}
    & {0.907${\scriptscriptstyle \pm .13}$} & {1.591${\scriptscriptstyle \pm .85}$} 
    & {$0.805{\scriptscriptstyle \pm.07}$} & {$0.882{\scriptscriptstyle \pm.08}$} 
    & {0.322${\scriptscriptstyle \pm.00}$} & {0.940${\scriptscriptstyle \pm.00}$}
    & {0.385${\scriptscriptstyle \pm.01}$} & {1.283${\scriptscriptstyle \pm.09}$} \\

\textsc{Segment-based} 
    & $0.709{\scriptscriptstyle \pm.06}$ & {1.464${\scriptscriptstyle \pm.09}$} 
    & {0.958${\scriptscriptstyle \pm.04}$} & {0.985${\scriptscriptstyle \pm.01}$} 
    & \textbf{0.121${\scriptscriptstyle \pm.01}$} & \textbf{0.982${\scriptscriptstyle \pm.00}$} 
    & {0.328${\scriptscriptstyle \pm.05}$} & {1.106${\scriptscriptstyle \pm.15}$}
     \\
\textsc{Fine-grain} 
    & \textbf{0.403${\scriptscriptstyle \pm.12}$} & \textbf{1.187${\scriptscriptstyle \pm.22}$} 
    & \textbf{0.960${\scriptscriptstyle \pm.02}$} & \textbf{0.992${\scriptscriptstyle \pm.00}$} 
    & {0.121${\scriptscriptstyle \pm.03}$} &{0.981${\scriptscriptstyle \pm.01}$} 
    & \textbf{0.130${\scriptscriptstyle \pm.02}$} & \textbf{0.926${\scriptscriptstyle \pm.19}$}
\\
\bottomrule
\end{tabular}
}
\end{table*}

\textbf{The Effect of Modality Combination.} Table~\ref{tab:ablation-results} evaluates how combining each modality impacts downstream performance. Relying solely on image features yields the weakest results across all benchmarks, showing that visual appearance alone is insufficient for road-level tasks without geographic or behavioral grounding. In comparison, geographic coordinates alone establish a strong spatial prior for road popularity and HBE due to spatial autocorrelation across neighboring areas, but are insufficient for dynamic tasks like traffic speed estimation. Introducing trajectory dynamics directly addresses this gap: pairing trajectories with coordinates (\textsc{Loc} + \textsc{Traj}) substantially outperforms the static pair (\textsc{Loc} + \textsc{SVI}) on traffic speed and reduces HBE error fivefold. Ultimately, integrating all three data sources in \model{} achieves the best overall results across every benchmark, demonstrating that they provide complementary signals, essential for learning general-purpose location representations.

\begin{table*}[t]
\centering
\caption{Modality combination ablation on San Francisco. \textbf{Best} and \underline{second best} values highlighted.}
\label{tab:ablation-results}
\footnotesize
\setlength{\tabcolsep}{2.5pt}
\resizebox{0.85\textwidth}{!}{%
\begin{tabular}{ccc cc cc cc cc}
\toprule
\multicolumn{3}{c}{\textbf{Modalities}} & \multicolumn{2}{c}{\textit{Traffic Speed}} & \multicolumn{2}{c}{\textit{Road Popularity}} & \multicolumn{2}{c}{\textit{AOI Function}} & \multicolumn{2}{c}{\textit{HBE}} \\
\cmidrule(lr){1-3} \cmidrule(lr){4-5} \cmidrule(lr){6-7} \cmidrule(lr){8-9} \cmidrule(lr){10-11}
\textbf{Loc} & \textbf{SVI} & \textbf{Traj} & ($\downarrow$) MAE & ($\downarrow$) RMSE & ($\uparrow$) F1 & ($\uparrow$) AUPRC & ($\downarrow$) L1 & ($\uparrow$) Cosine & ($\downarrow$) MAE & ($\downarrow$) RMSE \\
\midrule
\checkmark & -- & -- 
    & 1.117{\scriptsize $\pm .03$} & 2.101{\scriptsize $\pm .07$} 
    & 0.928{\scriptsize $\pm .03$} & 0.982{\scriptsize $\pm .00$} 
    & \underline{0.188}{\scriptsize $\pm .06$} & 0.956{\scriptsize $\pm .01$} 
    & 0.237{\scriptsize $\pm .05$} & 1.277{\scriptsize $\pm .12$} \\
-- & \checkmark & -- 
    & 2.203{\scriptsize $\pm .09$} & 4.384{\scriptsize $\pm .04$} 
    & 0.582{\scriptsize $\pm .06$} & 0.894{\scriptsize $\pm .02$} 
    & 1.114{\scriptsize $\pm .01$} & 0.648{\scriptsize $\pm .00$} 
    & 1.744{\scriptsize $\pm .04$} & 5.061{\scriptsize $\pm .11$} \\
\checkmark & \checkmark & -- 
    & 1.051{\scriptsize $\pm .08$} & \underline{1.724}{\scriptsize $\pm .17$} 
    & 0.921{\scriptsize $\pm .04$} & 0.972{\scriptsize $\pm .02$} 
    & 0.262{\scriptsize $\pm .01$} & 0.962{\scriptsize $\pm .00$} 
    & 1.076{\scriptsize $\pm .13$} & 1.653{\scriptsize $\pm .10$} \\
\checkmark & -- & \checkmark 
    & \underline{0.907}{\scriptsize $\pm .10$} & 1.774{\scriptsize $\pm .20$} 
    & \underline{0.952}{\scriptsize $\pm .00$} & \underline{0.984}{\scriptsize $\pm .00$} 
    & 0.204{\scriptsize $\pm .02$} & \underline{0.967}{\scriptsize $\pm .01$} 
    & \underline{0.191}{\scriptsize $\pm .00$} & \underline{1.128}{\scriptsize $\pm .02$} \\
\midrule
\checkmark & \checkmark & \checkmark 
    & \textbf{0.403}{\scriptsize $\pm .12$} & \textbf{1.187}{\scriptsize $\pm .22$} 
    & \textbf{0.960}{\scriptsize $\pm .02$} & \textbf{0.992}{\scriptsize $\pm .00$} 
    & \textbf{0.121}{\scriptsize $\pm .03$} & \textbf{0.981}{\scriptsize $\pm .01$} 
    & \textbf{0.130}{\scriptsize $\pm .02$} & \textbf{0.926}{\scriptsize $\pm .19$} \\
\bottomrule
\end{tabular}
}
%\vspace{-3mm}
\end{table*}

\textbf{Generalization to off-network locations.} A natural question is whether our framework can represent arbitrary locations that do not lie along observed trajectories. To evaluate this, we test our pretrained location encoder against baseline location backbones on an additional Point-of-Interest (POI) popularity prediction task in San Francisco, predicting venue popularity strictly from raw coordinates that lie off the drivable road network and away from observed
trajectories. As shown in Table~\ref{tab:poi-generalization}, \model{}'s location-only encoder outperforms the strongest baseline, with relative improvements of $10.6\%$ in accuracy, $28.3\%$ in macro-F1, and $3.9\%$ in AUROC. Absolute performance is modest on this task indicating that predicting venue popularity from coordinates alone is difficult for all models. The consistent ordering nonetheless suggests that \model{}'s location encoder to effectively generalizes to off-network coordinates rather than memorizing locations seen during pretraining.

\begin{table*}[t]
\centering
\caption{Off-network location-only performance on the POI popularity prediction task in San Francisco. \textbf{Best} values highlighted.}
\label{tab:poi-generalization}
\footnotesize
\setlength{\tabcolsep}{2pt} % Re-introduce some spacing for readability
\resizebox{0.55\textwidth}{!}{% This forces the table to fit the page width perfectly
\begin{tabular}{lccc}
\toprule
\textbf{Model} \textit{(location encoder only)} & $\uparrow$ {Accuracy} & $\uparrow$ {F1} & $\uparrow$ {AUROC} \\
\midrule
\textsc{SatCLIP} & 0.251 {\scriptsize $\pm$ .00} & 0.101 {\scriptsize $\pm$ .00} & 0.500 {\scriptsize $\pm$ .02} \\
\textsc{GeoCLIP} & 0.298 {\scriptsize $\pm$ .01} & 0.209 {\scriptsize $\pm$ .00} & 0.545 {\scriptsize $\pm$ .02} \\
\textsc{GAIR} & {0.310} {\scriptsize $\pm$ .02} & {0.247} {\scriptsize $\pm$ .03} & {0.568} {\scriptsize $\pm$ .01} \\
\midrule
\textbf{\model{}} & \textbf{0.343} {\scriptsize $\pm$ .01} & \textbf{0.317} {\scriptsize $\pm$ .02} & \textbf{0.590} {\scriptsize $\pm$ .01} \\
\bottomrule
\end{tabular}
}
%\vspace{-3mm}
\end{table*}

\section{Related Work}
\textbf{Trajectory Representation Learning.} Trajectory representation learning uses mobility traces as a primary signal to capture human travel dynamics across urban spaces. While existing trajectory foundation models successfully power downstream applications such as next-location prediction, classification, and trajectory generation~\citep{rao2020lstm, chen2021trajvae, zhu2023difftraj, zhu2025unitraj, li2017diffusion, anastasiou2022generation, zhou2026gps, yang2025bert4traj}, they almost universally formulate trajectories as discrete sequences of GPS points or spatial grid cells. Consequently, learned representations remain restricted to global sequence embeddings or discrete point- and cell-level tokens. This discrete formulation cannot be directly applied in our multimodal setting, where continuous trajectories must be aligned with static observations at locations that lie within or near a transit path without matching any GPS point. In such scenarios, whole-trajectory embeddings are too coarse, while discrete point tokens provide incomplete alignment anchors. In contrast, \model{}'s continuous trajectory representation via its neural implicit function enables querying localized embeddings at arbitrary target locations, unlocking fine-grained geospatial alignment.

\textbf{Geospatial Multimodal SSL.} Multimodal self-supervised learning has emerged as a prominent paradigm for pretraining geospatial foundation models, primarily aligning street-view, satellite images, and text with geographic coordinates~\citep{cepeda2023geoclip, klemmer2025satclip, mai2023csp, liu2024remoteclip, dhakal2025range, sastry2025taxabind}. While straightforward to co-locate, these static modalities only capture the visual appearance and semantics of locations, overlooking their dynamic urban function. Temporal dynamics have been explored in weather forecasting~\citep{nguyen2023climax, bi2023Pangu-Weather} and multi-temporal Earth observation embedding fields~\citep{brown2025alphaearth}. However, these approaches model how the physical environment evolves on fixed grids, rather than how humans navigate through it. Recently, \textsc{ME-POIs}~\citep{siampou2026mobility} introduced the concept of enriching static POI representations with human mobility signals, however, it models discrete check-in sequences rather than continuous trajectories. In contrast, \model{} is the first framework to incorporate continious trajectories into geospatial multimodal pretraining. By modeling trajectories via a continuous neural implicit function, \model{} extracts movement representations at arbitrary locations, enabling fine-grained alignment between street-view images and trajectories.

\section{Conclusion} \label{sec:conclusion}
We introduced \textsc{\model}, a trajectory-centric MSSL framework that aligns human mobility with street-view images and geographic locations at the finest granularity. By representing each trajectory as a path-conditioned neural implicit function, \textsc{\model} infers localized trajectory embeddings directly at street-view image coordinates and aligns them with the corresponding visual and location representations. Experiments showed that \textsc{\model} outperforms static multimodal baseslines and a trajectory foundation model on every task, with ablations attributing these gains the addition of trajectory modality and the alignment granularity. Future work will extend \model{} to additional modalities (e.g., road networks, building footprints, remote sensing imagery), scale pretraining from regional to global coverage, and explore zero- and few-shot transfer techniques.

%\newpage

\subsection*{AI use statement}

%(This section is \textbf{required} and does not count toward the page limit.)
In this work, we used generative AI tools for writing assistance (revising sections to improve clarity),  and for coding assistance during the implementation of our method. We have not used generative AI tools for research ideation or experimental design. All AI-assisted text and technical claim was read, edited, and approved by the authors. AI-assisted code was reviewed by the authors and validated through the experiments reported in this paper. We take responsibility for the final content of this work, including text, claims, or artifacts produced with the aid of generative AI.

\subsection*{Ethics statement}

%(This section is \textbf{recommended} and does not count toward the page limit.)

This paper presents work whose goal is to advance the field of Machine Learning. There are many potential societal consequences of our work, none of which we feel must be specifically highlighted here. Our multimodal, trajectory-centric geospatial framework could enable better location based services and urban planning tools. Our experiments rely on anonymized trajectory data and aggregated, anonymized tasks labels minimizing potential privacy concerns.

\subsection*{Reproducibility statement}

We have taken several steps to support the reproducibility of our work. The architecture of \model{}, is described in detail, and all hyperparameters, model sizes, and training details are listed in Appendix~\ref{apdx:experiments}. We also describe the downstream tasks, label definitions, discretization procedures, data splits, evaluation metrics, and the baseline fine-tuning protocol. The trajectory datasets and street-view images used for pretraining are publicly available and widely adopted by the geospatial research community. The downstream labels are derived from aggregated and anonymized proprietary data and cannot be released. We therefore document their construction in detail so that the evaluation can be replicated with comparable data sources. All results are averaged over three splits with standard deviations reported. Our codebase will be publicly available upon acceptance.

% \subsubsection*{Author Contributions}
% If you'd like to, you may include  a section for author contributions as is done
% in many journals. This is optional and at the discretion of the authors.

% \subsubsection*{Acknowledgments}
% Use unnumbered third level headings for the acknowledgments. All
% acknowledgments, including those to funding agencies, go at the end of the paper.

\bibliography{iclr2027_conference}

@article{liu2026gair,
  title={GAIR: Location-aware self-supervised contrastive pre-training with geo-aligned implicit representations},
  author={Liu, Zeping and Ni, Lao and Wang, Zhangyu and Jiao, Junfeng and Mai, Gengchen},
  journal={ISPRS Journal of Photogrammetry and Remote Sensing},
  page={166-182},
  volumn={237},
  year={2026}
}

@article{fuller2024croma,
  title={CROMA: Remote sensing representations with contrastive radar-optical masked autoencoders},
  author={Fuller, Anthony and Millard, Koreen and Green, James},
  journal={Advances in Neural Information Processing Systems},
  volume={36},
  year={2024}
}

@article{huang2023learning,
  title={Learning urban region representations with POIs and hierarchical graph infomax},
  author={Huang, Weiming and Zhang, Daokun and Mai, Gengchen and Guo, Xu and Cui, Lizhen},
  journal={ISPRS Journal of Photogrammetry and Remote Sensing},
  volume={196},
  pages={134--145},
  year={2023},
  publisher={Elsevier}
}

@article{liu2024remoteclip,
  title={Remoteclip: A vision language foundation model for remote sensing},
  author={Liu, Fan and Chen, Delong and Guan, Zhangqingyun and Zhou, Xiaocong and Zhu, Jiale and Ye, Qiaolin and Fu, Liyong and Zhou, Jun},
  journal={IEEE Transactions on Geoscience and Remote Sensing},
  year={2024},
  publisher={IEEE}
}

@article{bai2025integrating,
  title={Integrating remote sensing with OpenStreetMap data for comprehensive scene understanding through multi-modal self-supervised learning},
  author={Bai, Lubin and Zhang, Xiuyuan and Wang, Haoyu and Du, Shihong},
  journal={Remote Sensing of Environment},
  volume={318},
  pages={114573},
  year={2025},
  publisher={Elsevier}
}

@article{bi2023Pangu-Weather,
  title={Accurate medium-range global weather forecasting with 3D neural networks},
  author={Bi, Kaifeng and Xie, Lingxi and Zhang, Hengheng and Chen, Xin and Gu, Xiaotao and Tian, Qi},
  journal={Nature},
  volume={619},
  number={7970},
  pages={533--538},
  year={2023},
  publisher={Nature Publishing Group UK London}
}

@inproceedings{klemmer2025satclip,
  title={Satclip: Global, general-purpose location embeddings with satellite imagery},
  author={Klemmer, Konstantin and Rolf, Esther and Robinson, Caleb and Mackey, Lester and Ru{\ss}wurm, Marc},
  booktitle={Proceedings of the AAAI Conference on Artificial Intelligence},
  volume={39},
  number={4},
  pages={4347--4355},
  year={2025}
}

@inproceedings{zhu2025unitraj,
  title={UniTraj: Learning a Universal Trajectory Foundation Model from Billion-Scale Worldwide Traces},
  author={Zhu, Yuanshao and Yu, James Jianqiao and Zhao, Xiangyu and Zhou, Xun and Han, Liang and Wei, Xuetao and Liang, Yuxuan},
  booktitle={The Thirty-ninth Annual Conference on Neural Information Processing Systems},
  year={2025}
}

@article{siampou2025poly2vec,
  title={Poly2Vec: Polymorphic Fourier-Based Encoding of Geospatial Objects for GeoAI Applications},
  author={Siampou, Maria Despoina and Li, Jialiang and Krumm, John and Shahabi, Cyrus and Lu, Hua},
  journal={Proceedings of Machine Learning Research},
  volume={267},
  pages={55511--55532},
  year={2025}
}

@inproceedings{sastry2025taxabind,
  title={Taxabind: A unified embedding space for ecological applications},
  author={Sastry, Srikumar and Khanal, Subash and Dhakal, Aayush and Ahmad, Adeel and Jacobs, Nathan},
  booktitle={2025 IEEE/CVF Winter Conference on Applications of Computer Vision (WACV)},
  pages={1765--1774},
  year={2025},
  organization={IEEE}
}

@inproceedings{dhakal2025range,
  title={RANGE: Retrieval augmented neural fields for multi-resolution geo-embeddings},
  author={Dhakal, Aayush and Sastry, Srikumar and Khanal, Subash and Ahmad, Adeel and Xing, Eric and Jacobs, Nathan},
  booktitle={Proceedings of the IEEE/CVF Conference on Computer Vision and Pattern Recognition},
  pages={24680--24689},
  year={2025}
}

@article{mai2022review,
  title={A review of location encoding for GeoAI: methods and applications},
  author={Mai, Gengchen and others},
  journal={IJGIS},
  volume={36},
  number={4},
  pages={639--673},
  year={2022},
  publisher={Taylor \& Francis}
}

@article{mai2024opportunities,
  title={On the opportunities and challenges of foundation models for geoai (vision paper)},
  author={Mai, Gengchen and Huang, Weiming and Sun, Jin and Song, Suhang and Mishra, Deepak and Liu, Ninghao and Gao, Song and Liu, Tianming and Cong, Gao and Hu, Yingjie and others},
  journal={ACM Transactions on Spatial Algorithms and Systems},
  year={2024},
  publisher={ACM New York, NY}
}

@inproceedings{guo2024skysense,
  title={Skysense: A multi-modal remote sensing foundation model towards universal interpretation for earth observation imagery},
  author={Guo, Xin and Lao, Jiangwei and Dang, Bo and Zhang, Yingying and Yu, Lei and Ru, Lixiang and Zhong, Liheng and Huang, Ziyuan and Wu, Kang and Hu, Dingxiang and others},
  booktitle={Proceedings of the IEEE/CVF Conference on Computer Vision and Pattern Recognition},
  pages={27672--27683},
  year={2024}
}

@inproceedings{nguyen2023climax,
  title={ClimaX: A foundation model for weather and climate},
  author={Nguyen, Tung and Brandstetter, Johannes and Kapoor, Ashish and Gupta, Jayesh K and Grover, Aditya},
  booktitle={International Conference on Machine Learning},
  pages={25904--25938},
  year={2023},
  organization={PMLR}
}

@article{balsebre2023cityfm,
  title={Cityfm: City foundation models to solve urban challenges},
  author={Balsebre, Pasquale and Huang, Weiming and Cong, Gao and Li, Yi},
  journal={arXiv preprint arXiv:2310.00583},
  year={2023}
}

@inproceedings{rao2020lstm,
  title={LSTM-TrajGAN: A Deep Learning Approach to Trajectory Privacy Protection},
  author={Rao, Jinmeng and Gao, Song and Kang, Yuhao and Huang, Qunying},
  booktitle={11th International Conference on Geographic Information Science (GIScience 2021)-Part I (2020)},
  year={2020},
  organization={Schloss-Dagstuhl-Leibniz Zentrum f{\"u}r Informatik}
}

@article{vaswani2017attention,
  title={Attention is all you need},
  author={Vaswani, Ashish  and  Shazeer, Noam and  Parmar, Niki and  Uszkoreit, Jakob and  Jones, Llion and  Gomez, Aidan N. and  Kaiser, Lukasz and  Polosukhin, Illia},
  journal={Advances in Neural Information Processing Systems},
  year={2017}
}

@inproceedings{mai2020space2vec,
title={Multi-Scale Representation Learning for Spatial Feature Distributions using Grid Cells},
author={Mai, Gengchen and Janowicz, Krzysztof and Yan, Bo and Zhu, Rui and Cai, Ling and Lao, Ni},
booktitle={ICLR 2020},
year={2020},
organization={openreview}
}

@inproceedings{mai2023csp,
  title={CSP: Self-Supervised Contrastive Spatial Pre-Training for Geospatial-Visual Representations},
  author={Mai, Gengchen and Lao, Ni and He, Yutong and Song, Jiaming and Ermon, Stefano},
  booktitle={International Conference on Machine Learning},
  year={2023},
  organization={PMLR}
}

@inproceedings{cepeda2023geoclip,
  title={GeoCLIP: Clip-Inspired Alignment between Locations and Images for Effective Worldwide Geo-localization},
  author={Cepeda, Vicente Vivanco and Nayak, Gaurav Kumar and Shah, Mubarak},
  booktitle={Thirty-seventh Conference on Neural Information Processing Systems},
  year={2023}
}

@String{Computing = "Computing" }

@String{Computer = "{IEEE} Computer" }

@ArtifactSoftware{R,
    title = {R: A Language and Environment for Statistical Computing},
    author = {{R Core Team}},
    organization = {R Foundation for Statistical Computing},
    address = {Vienna, Austria},
    year = {2019},
    url = {https://www.R-project.org/},
}

@article{wang2018learning,
  title={Learning Discriminative Video Representations Using Adversarial Perturbations},
  author={Wang, Jue and Cherian, Anoop},
  journal={arXiv preprint arXiv:1807.09380},
  year={2018}
}

@inproceedings{chen2020simple,
  title={A simple framework for contrastive learning of visual representations},
  author={Chen, Ting and Kornblith, Simon and Norouzi, Mohammad and Hinton, Geoffrey},
  booktitle={International conference on machine learning},
  pages={1597--1607},
  year={2020},
  organization={PMLR}
}

@inproceedings{muhlematter2026urbanfusion,
    title={UrbanFusion: Stochastic Multimodal Fusion for Contrastive Learning of Robust Spatial Representations},
    author={Dominik J. M{\"u}hlematter and Lin Che and Ye Hong and Martin Raubal and Nina Wiedemann},
    booktitle={Forty-third International Conference on Machine Learning},
    year={2026},
    url={https://openreview.net/forum?id=C0NCLB6ZGY}
}

@article{zheng2014urban,
  title={Urban computing: concepts, methodologies, and applications},
  author={Zheng, Yu and Capra, Licia and Wolfson, Ouri and Yang, Hai},
  journal={ACM Transactions on Intelligent Systems and Technology (TIST)},
  volume={5},
  number={3},
  pages={1--55},
  year={2014},
  publisher={ACM New York, NY, USA}
}

@article{zhu2023difftraj,
  title={Difftraj: Generating gps trajectory with diffusion probabilistic model},
  author={Zhu, Yuanshao and Ye, Yongchao and Zhang, Shiyao and Zhao, Xiangyu and Yu, James},
  journal={Advances in Neural Information Processing Systems},
  volume={36},
  pages={65168--65188},
  year={2023}
}

@article{chen2021trajvae,
  title={TrajVAE: A Variational AutoEncoder model for trajectory generation},
  author={Chen, Xinyu and Xu, Jiajie and Zhou, Rui and Chen, Wei and Fang, Junhua and Liu, Chengfei},
  journal={Neurocomputing},
  volume={428},
  pages={332--339},
  year={2021},
  publisher={Elsevier}
}

@article{janowicz2025geofm,
  title={GeoFM: how will geo-foundation models reshape spatial data science and GeoAI?},
  author={Janowicz, Krzysztof and Mai, Gengchen and Huang, Weiming and Zhu, Rui and Lao, Ni and Cai, Ling},
  journal={International Journal of Geographical Information Science},
  volume={39},
  number={9},
  pages={1849--1865},
  year={2025},
  publisher={Taylor \& Francis}
}

@inproceedings{siampou2025toward,
  title={Toward Foundation Models for Mobility Enriched Geospatially Embedded Objects},
  author={Siampou, Maria Despoina and Hsu, Shang-Ling and Choudhury, Shushman and Arora, Neha and Shahabi, Cyrus},
  booktitle={Proceedings of the 33rd ACM International Conference on Advances in Geographic Information Systems},
  pages={774--779},
  year={2025}
}

@article{o2015ecml,
  title={Ecml/pkdd 15: Taxi trajectory prediction (i)},
  author={O’Connell, Meghan and Moreira-Matias, L and Kan, Wendy},
  journal={Kaggle. Retrieved April},
  volume={11},
  pages={2025},
  year={2015}
}

@article{li2026lagging,
  title={From Lagging to Leading: Validating Hard Braking Events as High-Density Indicators of Segment Crash Risk},
  author={Li, Yechen and Shahane, Shantanu and Vasserman, Shoshana and Osorio, Carolina and Chen, Yi-fan and Kuznetsov, Ivan and White, Kristin and Swiatkowska, Justyna and Arora, Neha and Guo, Feng},
  journal={arXiv preprint arXiv:2601.06327},
  year={2026}
}

@article{siampou2026mobility,
  title={Mobility-Embedded POIs: Learning What A Place Is and How It Is Used from Human Movement},
  author={Siampou, Maria Despoina and Choudhury, Shushman and Hsu, Shang-Ling and Arora, Neha and Shahabi, Cyrus},
  journal={arXiv preprint arXiv:2601.21149},
  year={2026}
}

@article{zhou2024road,
  title={Road network representation learning with the third law of geography},
  author={Zhou, Haicang and Huang, Weiming and Chen, Yile and He, Tiantian and Cong, Gao and Ong, Yew-Soon},
  journal={Advances in Neural Information Processing Systems},
  volume={37},
  pages={11789--11813},
  year={2024}
}

@article{brown2025alphaearth,
  title={Alphaearth foundations: An embedding field model for accurate and efficient global mapping from sparse label data},
  author={Brown, Christopher F and Kazmierski, Michal R and Pasquarella, Valerie J and Rucklidge, William J and Samsikova, Masha and Zhang, Chenhui and Shelhamer, Evan and Lahera, Estefania and Wiles, Olivia and Ilyushchenko, Simon and others},
  journal={arXiv preprint arXiv:2507.22291},
  year={2025}
}

@inproceedings{choudhury2024towards,
  title={Towards a trajectory-powered foundation model of mobility},
  author={Choudhury, Shushman and Kreidieh, Abdul Rahman and Kuznetsov, Ivan and Arora, Neha},
  booktitle={Proceedings of the 3rd ACM SIGSPATIAL International Workshop on Spatial Big Data and AI for Industrial Applications},
  pages={1--4},
  year={2024}
}

@article{choudhury2025s2vec,
  title={S2Vec: Self-Supervised Geospatial Embeddings for the Built Environment},
  author={Choudhury, Shushman and Suvarna, Chandrakumari and Tsogsuren, Iveel and Kreidieh, Abdul Rahman and Aharoni, Elad and Lu, Chun-Ta and Arora, Neha},
  journal={ACM Transactions on Spatial Algorithms and Systems},
  year={2025},
  publisher={ACM New York, NY}
}

@article{dosovitskiy2020image,
  title={An image is worth 16x16 words: Transformers for image recognition at scale},
  author={Dosovitskiy, Alexey and Beyer, Lucas and Kolesnikov, Alexander and Weissenborn, Dirk and Zhai, Xiaohua and Unterthiner, Thomas and Dehghani, Mostafa and Minderer, Matthias and Heigold, Georg and Gelly, Sylvain and others},
  journal={arXiv preprint arXiv:2010.11929},
  year={2020}
}

@article{kazemi2019time2vec,
  title={Time2vec: Learning a vector representation of time},
  author={Kazemi, Seyed Mehran and Goel, Rishab and Eghbali, Sepehr and Ramanan, Janahan and Sahota, Jaspreet and Thakur, Sanjay and Wu, Stella and Smyth, Cathal and Poupart, Pascal and Brubaker, Marcus},
  journal={arXiv preprint arXiv:1907.05321},
  year={2019}
}

@article{oord2018representation,
  title={Representation learning with contrastive predictive coding},
  author={Oord, Aaron van den and Li, Yazhe and Vinyals, Oriol},
  journal={arXiv preprint arXiv:1807.03748},
  year={2018}
}

@inproceedings{zhao2025unitr,
  title={UniTR: A Unified Framework for Joint Representation Learning of Trajectories and Road Networks},
  author={Zhao, Jie and Chen, Chao and Zhu, Yuanshao and Deng, Mingyu and Liang, Yuxuan},
  booktitle={Proceedings of the AAAI Conference on Artificial Intelligence},
  volume={39},
  number={12},
  pages={13348--13356},
  year={2025}
}

@inproceedings{he2020momentum,
  title={Momentum contrast for unsupervised visual representation learning},
  author={He, Kaiming and Fan, Haoqi and Wu, Yuxin and Xie, Saining and Girshick, Ross},
  booktitle={Proceedings of the IEEE/CVF conference on computer vision and pattern recognition},
  pages={9729--9738},
  year={2020}
}

@inproceedings{yang2025bert4traj,
  title={BERT4Traj: Transformer-Based Trajectory Reconstruction for Sparse Mobility Data},
  author={Yang, Hao and Yao, Angela and Whalen, Christopher C and Mai, Gengchen},
  booktitle={13th International Conference on Geographic Information Science (GIScience 2025)},
  pages={8--1},
  year={2025},
  organization={Schloss Dagstuhl--Leibniz-Zentrum f{\"u}r Informatik}
}

@misc{piorkowski2009crawdad,
  title={CRAWDAD data set epfl/mobility (v. 2009-02-24)},
  author={Piorkowski, Michal and Sarafijanovic-Djukic, Natasa and Grossglauser, Matthias},
  year={2009}
}

@article{yang2018fast,
  title={Fast map matching, an algorithm integrating hidden Markov model with precomputation},
  author={Yang, Can and Gidofalvi, Gyozo},
  journal={International Journal of Geographical Information Science},
  volume={32},
  number={3},
  pages={547--570},
  year={2018},
  publisher={Taylor \& Francis}
}

@article{li2017diffusion,
  title={Diffusion convolutional recurrent neural network: Data-driven traffic forecasting},
  author={Li, Yaguang and Yu, Rose and Shahabi, Cyrus and Liu, Yan},
  journal={arXiv preprint arXiv:1707.01926},
  year={2017}
}

@inproceedings{anastasiou2022generation,
  title={Generation of synthetic urban vehicle trajectories},
  author={Anastasiou, Chrysovalantis and Kim, Seon Ho and Shahabi, Cyrus},
  booktitle={2022 IEEE International Conference on Big Data (Big Data)},
  pages={359--366},
  year={2022},
  organization={IEEE}
}

@article{zhou2026gps,
  title={From GPS Points to Travel Patterns: Flexible and Semantic Trajectory Generation with LLMs},
  author={Zhou, Silin and Wang, Chenhao and Wen, Yuntao and Shang, Shuo and Chen, Lisi and Kalnis, Panos},
  journal={arXiv preprint arXiv:2605.30014},
  year={2026}
}

@inproceedings{muhlematter2025urbanfusion,
  title={Urbanfusion: Stochastic multimodal fusion for contrastive learning of robust spatial representations},
  author={M{\"u}hlematter, Dominik J and Che, Lin and Hong, Ye and Raubal, Martin and Wiedemann, Nina},
  booktitle={Forty-third International Conference on Machine Learning},
  year={2025}
}

@article{xiao2025foundation,
  title={Foundation models for remote sensing and earth observation: A survey},
  author={Xiao, Aoran and Xuan, Weihao and Wang, Junjue and Huang, Jiaxing and Tao, Dacheng and Lu, Shijian and Yokoya, Naoto},
  journal={IEEE Geoscience and Remote Sensing Magazine},
  year={2025},
  publisher={IEEE}
}

@inproceedings{wang2025gtr,
  title={GTR: A general, multi-view, and dynamic framework for trajectory representation learning},
  author={Wang, Xiangheng and Fang, Ziquan and Huang, Chenglong and Hu, Danlei and Chen, Lu and Gao, Yunjun},
  booktitle={Forty-second International Conference on Machine Learning},
  year={2025}
}

@inproceedings{ma2024more,
  title={More than routing: Joint GPS and route modeling for refine trajectory representation learning},
  author={Ma, Zhipeng and Tu, Zheyan and Chen, Xinhai and Zhang, Yan and Xia, Deguo and Zhou, Guyue and Chen, Yilun and Zheng, Yu and Gong, Jiangtao},
  booktitle={Proceedings of the ACM Web Conference 2024},
  pages={3064--3075},
  year={2024}
}

@inproceedings{li2018deep,
  title={Deep representation learning for trajectory similarity computation},
  author={Li, Xiucheng and Zhao, Kaiqi and Cong, Gao and Jensen, Christian S and Wei, Wei},
  booktitle={2018 IEEE 34th international conference on data engineering (ICDE)},
  pages={617--628},
  year={2018},
  organization={IEEE}
}

@inproceedings{zhou2025blurred,
  title={Blurred encoding for trajectory representation learning},
  author={Zhou, Silin and Chen, Yao and Shang, Shuo and Chen, Lisi and He, Bingsheng and Shibasaki, Ryosuke},
  booktitle={Proceedings of the 31st ACM SIGKDD Conference on Knowledge Discovery and Data Mining V. 2},
  pages={4132--4143},
  year={2025}
}

@article{lin2024unite,
  title={UniTE: a survey and unified pipeline for pre-training spatiotemporal trajectory embeddings},
  author={Lin, Yan and Zhou, Zeyu and Liu, Yicheng and Lv, Haochen and Wen, Haomin and Li, Tianyi and Li, Yushuai and Jensen, Christian S and Guo, Shengnan and Lin, Youfang and others},
  journal={IEEE Transactions on Knowledge and Data Engineering},
  volume={37},
  number={3},
  pages={1475--1494},
  year={2024},
  publisher={IEEE}
}

@article{lin2024trajfm,
  title={Trajfm: A vehicle trajectory foundation model for region and task transferability},
  author={Lin, Yan and Wei, Tonglong and Zhou, Zeyu and Wen, Haomin and Hu, Jilin and Guo, Shengnan and Lin, Youfang and Wan, Huaiyu},
  journal={arXiv preprint arXiv:2408.15251},
  year={2024}
}

@book{batty2013new,
  title={The new science of cities},
  author={Batty, Michael},
  year={2013},
  publisher={The MIT press}
}

@inproceedings{yuan2012discovering,
  title={Discovering regions of different functions in a city using human mobility and POIs},
  author={Yuan, Jing and Zheng, Yu and Xie, Xing},
  booktitle={Proceedings of the 18th ACM SIGKDD international conference on Knowledge discovery and data mining},
  pages={186--194},
  year={2012}
}

@article{hou2024global,
  title={Global Streetscapes—A comprehensive dataset of 10 million street-level images across 688 cities for urban science and analytics},
  author={Hou, Yujun and Quintana, Matias and Khomiakov, Maxim and Yap, Winston and Ouyang, Jiani and Ito, Koichi and Wang, Zeyu and Zhao, Tianhong and Biljecki, Filip},
  journal={ISPRS Journal of Photogrammetry and Remote Sensing},
  volume={215},
  pages={216--238},
  year={2024},
  publisher={Elsevier}
}

@article{liu2020visualizing,
  title={Visualizing and exploring POI configurations of urban regions on POI-type semantic space},
  author={Liu, Kang and Yin, Ling and Lu, Feng and Mou, Naixia},
  journal={Cities},
  volume={99},
  pages={102610},
  year={2020},
  publisher={Elsevier}
}
\bibliographystyle{iclr2027_conference}

\appendix
\newpage
\section{Appendix}

\subsection{Additional Details on Experimental Setup}
\label{apdx:experiments}

\subsubsection{Dataset Statistics and Preprocessing}
\label{apdx:dataset-stats}

\textbf{Street-View Images.} We download street-view images via the Mapillary API. To ensure visual fidelity, we filter out low-visibility and obstructed frames by excluding images with a \texttt{quality\_score} below $0.8$, using the automated quality metrics provided in the Mapillary metadata. Each retained image is associated with its camera capture coordinates and the identifier of its snapped OpenStreetMap (OSM)~\footnote{https://openstreetmap.org} road segment.

\textbf{Trajectory Preprocessing and Map Matching.} We spatially filter the raw trajectory datasets to retain only trips that fall within the geographic bounding box established by the street-view image dataset. Following standard practices in trajectory modeling~\citep{zhu2025unitraj, wang2018learning}, we extract the drivable road network for each study region from OSM using the \texttt{OSMnx} library. To correct for GPS noise, each raw trajectory point is map-matched to the road network graph~\citep{yang2018fast}. We filter out short trips with less than 10 points to ensure meaningful sequential context. Because both trajectories and street-view images are aligned to the same underlying OSM network, we use these shared segment identifiers to determine the query set $\mathcal{Q}_{\trajidx}$ for each trajectory $\trajseq_{\trajidx}$.

\textbf{Summary Statistics.} Table~\ref{tab:dataset_statistics} summarizes the statistics of our multimodal datasets for both regions.

\begin{table}[htbp]
\centering
\small
\caption{Summary statistics of the preprocessed multimodal datasets.}
\label{tab:dataset_statistics}
\begin{tabular}{ccc}
\toprule
\textbf{Metric} & \textbf{Porto} & \textbf{San Francisco} \\
\midrule
Bounding Box \texttt{(lat, lon)} & $[41.13, 41.19] \times [-8.69, -8.55]$ & $[37.74, 37.77] \times [-122.45, -122.43]$ \\
Time range & 05/17/2008 - 06/10/2008 &  07/01/2013 - 06/30/2014\\
\# Trajectories & $674{,}740$ & $5{,}381$ \\
\# Street-view images & $51{,}729$ & $68{,}040$ \\
\bottomrule
\end{tabular}
\end{table} 

\subsubsection{Downstream Tasks} \label{apdx:downstream_tasks}
We evaluate on four road-level tasks that are representative of fine-grained urban understanding. These tasks cover complementary aspects of a road segment: its traffic conditions (traffic speed), destination demand (road popularity), surrounding land use (AOI function), and safety risk (HBEs). To assess generalization beyond the road network, we also evaluate on the POI popularity task. All labels are defined per road segment. Each example is represented by the coordinate of a street-view image on the segment, which the model embeds directly.
\begin{itemize}[leftmargin=*]
    \item \textbf{Traffic speed}: A regression task that predicts the average observed traffic speed on each road segment (km/h), derived from aggregated and anonymized Google Maps traffic data that are independent of the input trajectories. Traffic speed can differ from the posted speed limit, as it reflects real-world driving behavior. There are 40,338 labels in San Francisco.
    
    \item \textbf{Road popularity}:We define road popularity as a proxy for user interest in reaching destinations along a road segment, measured by the average number of Google Maps direction queries to the segment. Since we do not observe actual visits, this provides an indirect signal of the segment's popularity as a destination. We discretize the continuous query values into four ordinal classes from low to high intent. This task evaluates the model's ability to predict destination demand, i.e., where people intend to go. There are 2,084 labels in San Francisco, and 1,752 in Porto.
    
    \item \textbf{Area-of-interest (AOI) function}: The AOI function prediction task aims to predict the distribution of POI categories associated with a specific road segment. To construct the ground truth, we define an AOI by generating a 50-meter buffer around each road segment and retrieving all enclosed POIs. We then compute and normalize the frequency of each POI category to create a probability distribution, which serves as a proxy for the segment's underlying function~\citep{liu2020visualizing, huang2023learning, zhou2024road}. This distribution is assigned as the ground truth label for all streetview images located along that segment, and is tested against the predicted distribution of our fine-tuned model~\citep{liu2020visualizing}. There are 56,549 labels in San Francisco, and 20,199 in Porto.
   
    \item \textbf{Hard breaking events (HBE)}: A regression task that predicts the number of hard-braking events on each road segment, derived from aggregated and anonymized data from the Google Android Auto platform. Hard-braking events indicate evasive maneuvers and have been validated as a segment-level surrogate for crash risk~\citep{li2026lagging}. There are 2,974 labels in San Francisco, and 1,282 in Porto.
    
    \item \textbf{POI popularity}: Similar to road-segment popularity, we define visit intent as a proxy for user interest in visiting a POI, measured by the average number of Google Maps direction queries to the location. Since we do not observe actual visits, this provides an indirect signal of interest. We discretize the continuous query values into four ordinal classes from low to high intent. In San Fransciso, we evaluate on 1,629 POIs.
\end{itemize}

\subsubsection{Details on Baselines} \label{apdx:baselines}

We compare \model{} against the latest geospatial MSSL models spanning the modalities it integrates: location encoders pretrained with visual supervision (\textsc{GeoCLIP}, \textsc{SatCLIP}), a static multimodal model aligning street-view imagery, remote sensing imagery, and coordinates (\textsc{GAIR}), and a trajectory foundation model (\textsc{UniTraj}). Each covers a subset of \model{}'s modalities.
\begin{itemize}[leftmargin=*]
    \item \textbf{\textsc{GeoCLIP}}~\citep{cepeda2023geoclip} is a CLIP-inspired geospatial model that contrastively aligns images with their GPS coordinates. GeoCLIP's image encoder builds on a frozen CLIP backbone, while its location encoder models the Earth as a continuous function, combining Random Fourier Feature (RFF) positional encodings at multiple resolutions into a hierarchical location representation. Although developed for worldwide image geolocalization, its pretrained location encoder has since been adopted as a general-purpose location representation for downstream geospatial prediction tasks~\citep{klemmer2025satclip, muhlematter2026urbanfusion}. We use its pretrained location encoder to embed each query coordinate.
    \item \textbf{\textsc{SatCLIP}}~\citep{klemmer2025satclip} is a global, general-purpose location encoder pretrained by contrastively matching openly available satellite imagery with its geographic coordinates. Its location encoder combines spherical harmonic basis functions with a sinusoidal representation network, yielding embeddings that summarize the environmental characteristics of any location on Earth. Similar to \textsc{GeoCLIP}, we use its pretrained location encoder to embed each query coordinate.
    \item \textbf{\textsc{GAIR}}~\citep{liu2026gair} is a location-aware self-supervised framework that jointly pretrains factorized encoders for overhead remote sensing imagery, street-view imagery, and geographic coordinates with contrastive objectives. To align street-view images with overhead imagery, it interpolates ViT patch features within each remote sensing image to obtain a localized overhead feature at the street-view location. Since GAIR has highlighted the predictive value of each of its encoders, we evaluate three variants: its location encoder \textsc{GAIR$_{\text{ loc}}$}, its street-view encoder \textsc{GAIR$_{\text{ SVI}}$}, and their combination \textsc{GAIR$_{\text{ loc+SVI}}$}. Our \model{} adopts \textsc{GAIR}'s street-view encoder as its visual backbone, so \textsc{GAIR$_{\text{ SVI}}$} corresponds to \model{} without the trajectory and location modality. \textsc{GAIR} is pretrained on Streetscapes1M, which samples one million street-view images across 688 cities from the Global Streetscapes dataset~\citep{hou2024global}. The street-view images used to pretrain \model{} are drawn from the same Global Streetscapes source.
    \item \textbf{\textsc{UniTraj}}~\citep{zhu2025unitraj} is a universal trajectory foundation model pretrained on WorldTrace~\footnote{https://huggingface.co/datasets/OpenTrace/WorldTrace}, a corpus of 2.45 million trajectories spanning 70 countries. It employs an encoder-decoder Transformer with rotary position encodings, trained with self-supervised trajectory masking and adaptive resampling to handle heterogeneous sampling rates. Notably, \textsc{UniTraj} has been evaluated on the Porto taxi dataset in both zero-shot and fine-tuning settings, so one of our two study regions is not foreign to it. \textsc{UniTraj} produces embeddings only at observed GPS points and, like other trajectory encoders, cannot be queried at coordinates where no point was recorded. We therefore obtain a representation for each road segment by mean-pooling the embeddings of all GPS points assigned to that segment, and fine-tune the model with the same task heads as \model{}.
\end{itemize}

\subsubsection{Implementation Details}
\label{apdx:implementation-details}

\textbf{Data Splits.} We partition the data using a spatial split over the road network. Road segments are divided into disjoint training and test sets such that approximately $60\%$ of trajectories fall in the training set and $40\%$ in the test set, with each trajectory assigned entirely to one side. Similarly, each street-view image is assigned to the same split as the road segment it is associated with. Consequently, evaluation query locations, along with their street-view images, labels, and intersecting trajectories, are unseen during model fine-tuning. Our \model{} and all baselines are fine-tuned on the same training split and evaluated on the test split. All reported results are averaged over three independent spatial splits, and we report mean and standard deviation across runs. 

\textbf{Coordinate Normalization.} Given our focus on specific urban regions across two datasets, we normalize all geographic coordinates into the range $[0, 1]$ relative to each area's bounding box. This local normalization ensures the model remains sensitive to fine-grained regional structures.

\textbf{Sampling and Augmentation.} During pretraining, we employ a trajectory-centric alignment strategy, where each sample consists of a single trajectory and $32$ street-view images whose geolocations lie along the trajectory's continuous spatial footprint. We enforce a fixed trajectory window size of ${\trajseqlen_{\trajidx}}=240$, applying random downsampling for longer trajectories and zero-padding for shorter ones. This stochastic dropping of waypoints serves as data augmentation and forces the model to learn trajectory representations invariant to varying sampling frequencies and diverse temporal gaps. Street-view images
undergo standard augmentations following prior work~\citep{cepeda2023geoclip, chen2020simple, liu2026gair}. During fine-tuning, we shift to a location-centric sampling strategy, where each sample consists of a street-view image, its associated location, and $16$ sampled intersecting trajectories.

\textbf{Architecture and Optimization.} We use a shared projection dimension of $128$ and a location embedding dimension of $128$. For the Space2Vec encoders, we set $\lambda_{\min}=1$, $\lambda_{\max}=\sqrt{2}$, and use $64$ spatial scales. The trajectory encoder is a $4$-layer Transformer with $8$ attention heads and feed-forward dimension $512$. We pretrain for $300$ epochs using Adam
with batch size $256$, learning rate $10^{-4}$, and weight decay $10^{-4}$, and maintain a location negative queue of $2048$ embeddings. For downstream fine-tuning, we train for $100$ epochs using a two-layer MLP prediction head with hidden dimension $1024$ and ReLU activation.

\textbf{Baselines.} We use the publicly released checkpoints for GeoCLIP and SatCLIP, and follow the hyperparameters provided in the official implementations of GAIR and UniTraj. All baselines are initialized from their released pretrained weights and fine-tuned end-to-end with the same task heads, folds, and training protocol as \model{}. Because the methods require different inputs, each is evaluated on the locations where its inputs are available.

\subsubsection{Pretraining Objective}
We utilize InfoNCE loss~\citep{oord2018representation} to pretrain our \model{}. In more detail, we define the pairwise InfoNCE loss between anchor $\mathbf{u}$ and positive target $\mathbf{v}^+$ over negative set $\mathcal{N}$ as:
\begin{equation*}
    \mathcal{L}_{\mathrm{info}}(\mathbf{u}, \mathbf{v}^+; \mathcal{N}) = -\log \frac{\exp\left(\mathrm{sim}(\mathbf{u}, \mathbf{v}^+) / \tau\right)}{\exp\left(\mathrm{sim}(\mathbf{u}, \mathbf{v}^+) / \tau\right) + \sum_{\mathbf{v}^- \in \mathcal{N}} \exp\left(\mathrm{sim}(\mathbf{u}, \mathbf{v}^-) / \tau\right)},
    \label{eq:infonce}
\end{equation*}
where $\mathrm{sim}(\cdot, \cdot)$ denotes cosine similarity scaled by temperature $\tau$.

\subsubsection{Computational Environment}
\label{apdx:computational-env}

Our experiments are performed on a cluster node equipped with 8 CPU cores, 128 GB of memory, and one NVIDIA H100 GPU. Furthermore, all neural network models are implemented in Python 3.10 using PyTorch 2.3.0 and CUDA 12.1.

\subsubsection{Licenses and Terms of Use}
\label{apdx:license}

We use publicly available trajectory datasets and street-view imagery sources. The Porto taxi trajectory dataset from the ECML/PKDD 2015 Taxi Trajectory Prediction Challenge is distributed under the Creative Commons Attribution 4.0 International license (CC BY 4.0), and we cite the original dataset source. The Cabspotting dataset contains taxi mobility traces from San Francisco and was originally provided by the Exploratorium~\cite{piorkowski2009crawdad}. Street-view images are retrieved from Mapillary, whose images are shared under a Creative Commons Attribution-ShareAlike license (CC BY-SA). We follow Mapillary's terms of use. All datasets are used explicitly for research purposes, with attribution to the original creators and providers.

%\subsection{Additional Details on Baselines}
\label{apdx:baselines}

\subsection{Evaluation of Query-Time Encoding Strategies}

\begin{table*}[!ht]
\centering
\caption{Ablation study on query-time temporal encoding strategies for the trajectory neural implicit function: synthetic-time query encoding and masked-time query encoding --  as we described in Section~\ref{sec:traj_nif}. 
The results of each setting on the SF dataset are averaged over 3 runs, and "${\scriptscriptstyle \pm X}$" denotes the standard deviation.
\textbf{Best} and \underline{second best} values are highlighted.}

\label{tab:time-results}
\footnotesize
\setlength{\tabcolsep}{2pt} % Re-introduce some spacing for readability
\resizebox{0.85\textwidth}{!}{% This forces the table to fit the page width perfectly
\begin{tabular}{l cc cc cc cc}
\toprule
\textbf{Method} & \multicolumn{2}{c}{\textit{Traffic Speed}} & \multicolumn{2}{c}{\textit{Road Popularity}} & \multicolumn{2}{c}{\textit{AOI Function}} & \multicolumn{2}{c}{\textit{HBE}} \\
\cmidrule(lr){2-3} \cmidrule(lr){4-5} \cmidrule(lr){6-7} \cmidrule(lr){8-9}
& ($\downarrow$) MAE & ($\downarrow$) RMSE & ($\uparrow$) F1 & ($\uparrow$) AUPRC & ($\downarrow$) L1 & ($\uparrow$) Cosine & ($\downarrow$) MAE & ($\downarrow$) RMSE \\
\midrule
\textsc{synthetic-time} 
    & \textbf{0.403${\scriptscriptstyle \pm.12}$} & \textbf{1.187${\scriptscriptstyle \pm.22}$} 
    & {0.960${\scriptscriptstyle \pm.02}$} & \textbf{0.992${\scriptscriptstyle \pm.00}$} 
    & \textbf{0.121${\scriptscriptstyle \pm.03}$} & \textbf{0.981${\scriptscriptstyle \pm.01}$} 
    & \textbf{0.130${\scriptscriptstyle \pm.02}$} & {0.926${\scriptscriptstyle \pm.19}$} \\
\textsc{masked-time} 
    & {0.451${\scriptscriptstyle \pm.04}$} & {1.230${\scriptscriptstyle \pm.07}$}
    & \textbf{0.970${\scriptscriptstyle \pm.08}$} & {0.987${\scriptscriptstyle \pm.01}$}
    & {0.179${\scriptscriptstyle \pm.04}$} & {0.972${\scriptscriptstyle \pm.01}$}
    & {0.147${\scriptscriptstyle \pm.02}$} & \textbf{0.873${\scriptscriptstyle \pm.20}$} \\
\bottomrule
\end{tabular}
}
\end{table*}

We empirically compare the two proposed query-time encoding strategies, to study their impact on downstream performance. Table~\ref{tab:time-results} presents the results. Overall, the synthetic-time variant achieves stronger performance across most metrics. This suggests that although street-view image locations are not observed GPS samples, the interpolated timestamp allows the model to place each query point within the continuous temporal path of the trajectory, which is especially useful for tasks that depend on localized mobility context. In contrast, the masked-time variant avoids the explicit assumption about the query timestamp, but requires the Transformer to infer the temporal position of the query token from the observed trajectory samples. This inference becomes more difficult when trajectories are sparse or irregularly sampled, which may explain its weaker overall performance. Nevertheless, the masked-time variant remains competitive and achieves the best result on a few metrics, suggesting that explicit temporal interpolation is not always necessary and may occasionally introduce noise. Taken together, these results suggest that synthetic-time encoding provides a strong overall design choice despite its simplicity, while masked-time encoding remains a reasonable alternative when interpolation is unreliable.

\end{document}